\documentclass{article}
\usepackage{microtype}
\usepackage{graphicx}
\usepackage{booktabs} 

\usepackage{hyperref}
\usepackage{caption}
\usepackage{subcaption}

\usepackage{multirow}
\usepackage[utf8]{inputenc}

\usepackage{listings} 
\usepackage{amssymb}
\usepackage{wrapfig}

\usepackage{etoolbox}
\BeforeBeginEnvironment{appendices}{\clearpage}

\usepackage{booktabs} 
\usepackage{algorithm} 
\usepackage[noend]{algpseudocode} 
\algdef{SE}[DOWHILE]{Do}{doWhile}{\algorithmicdo}[1]{\algorithmicwhile\ #1}%
\algblockdefx{ParForAll}{EndParFor}[1]%
  {\textbf{for all }#1 \textbf{do in parallel}}%
  {\textbf{end for}}
\algtext*{EndParFor}
\renewcommand{\algorithmicforall}{\textbf{for each}}

\algrenewcommand\textproc{}

\usepackage{soul}
\usepackage{graphicx}
\usepackage{color}
\usepackage{enumerate}
\usepackage{enumitem}
\usepackage{hyperref}
\usepackage{array}
\usepackage{listings}
\usepackage{verbatim}

\usepackage{sidecap} 
\usepackage{amsmath} 
\usepackage{amsfonts}
\usepackage{amssymb}
\usepackage{cleveref}

\usepackage[compact]{titlesec}
\newcommand{\rem}[1]{}

\renewcommand{\int}{\cap}
\newcommand{\define}{ \mathrel{\bf \colon \kern -2pt =} }	

\usepackage[compact]{titlesec}
    \titlespacing{\section}{0pt}{1ex}{1ex}
    \titlespacing{\subsection}{0pt}{1ex}{0ex}
    \titlespacing{\subsubsection}{0pt}{0.5ex}{0ex}

\usepackage{afterpage}

\usepackage[accepted]{mlsys2023}

\begin{document}
\afterpage{\cfoot{\thepage}}

\twocolumn[
\mlsystitle{Extreme Acceleration of Graph Neural Network-based Prediction Models for Quantum Chemistry}



\mlsyssetsymbol{equal}{*}

\begin{mlsysauthorlist}
\mlsysauthor{Hatem Helal}{gcc}
\mlsysauthor{Jesun Firoz}{ps}
\mlsysauthor{Jenna Bilbrey}{pr}
\mlsysauthor{Mario Michael Krell}{gcp}
\mlsysauthor{Tom Murray}{gcb}
\mlsysauthor{Ang Li}{pr}
\mlsysauthor{Sotiris Xantheas}{pr}
\mlsysauthor{Sutanay Choudhury}{pr}
\end{mlsysauthorlist}

\mlsysaffiliation{pr}{Pacific Northwest National Laboratory, Richland, USA}
\mlsysaffiliation{ps}{Pacific Northwest National Laboratory, Seattle, USA}
\mlsysaffiliation{gcc}{Graphcore, Cambridge, UK}
\mlsysaffiliation{gcp}{Graphcore, Palo Alto, USA}
\mlsysaffiliation{gcb}{Graphcore, Bristol, USA}

\mlsyscorrespondingauthor{Hatem Helal}{hatemh@graphcore.ai}
\mlsyscorrespondingauthor{Sutanay Choudhury}{sutanay.choudhury@pnnl.gov}

\mlsyskeywords{Graph neural networks, molecular}

\vskip 0.3in

\begin{abstract} 
Molecular property calculations are the bedrock of chemical physics.  High-fidelity \textit{ab initio} modeling techniques for computing the molecular properties can be prohibitively expensive, and motivate the development of machine-learning models that make the same predictions more efficiently.  Training graph neural networks over large molecular databases introduces unique computational challenges such as the need to process millions of small graphs with variable size and support communication patterns that are distinct from learning over large graphs such as social networks.   This paper demonstrates a novel hardware-software co-design approach to scale up the training of graph neural networks for molecular property prediction.  We introduce an algorithm to coalesce the batches of molecular graphs into fixed size packs to eliminate redundant computation and memory associated with alternative padding techniques and improve throughput via minimizing communication. We demonstrate the effectiveness of our co-design approach by providing an implementation of a well-established molecular property prediction model on the Graphcore Intelligence Processing Units (IPU).  We evaluate the training performance on multiple molecular graph databases with varying degrees of graph counts, sizes and sparsity.   We demonstrate that such a co-design approach can reduce the training time of such molecular property prediction models from days to less than two hours, opening new possibilities for AI-driven scientific discovery.
\end{abstract}

]



\printAffiliationsAndNotice{}  

\section{Introduction}
Recent advancements in the field of Artificial Intelligence (AI) have fueled notable progress in data-driven scientific discovery. Neural Networks (NNs) provide particular advantage as surrogate models for modeling quantum chemical properties \cite{DTNN2017}. High-accuracy \textit{ab initio} methods, which contain no empirical fitting parameters, are prohibitively expensive, with computational costs that scale as high as $O(N^7)$, where $N$ is the number of atoms \cite{kulichenko2021rise}. NNs are able to attain the same level of accuracy as the \textit{ab initio} technique on which the NN was trained but with roughly $O(N)$ scaling \cite{behler2017first, Smith2017, HIPNN}. Molecular structures are naturally represented as graphs, which makes Graph Neural Networks (GNNs) ~\cite{GCNConv, GATConv, SageConv} compelling for building surrogate models for quantum chemistry. Initially formulated as a message passing neural network \cite{gilmer2017neural}, adding support for both the bonding graph structure and atom-level attributes such as types and 3D coordinates became a key feature of such models \cite{SchNet2018}.  The past few years witnessed further expansion of this class of models, henceforth referred to as \textsl{molecular GNNs} through sophisticated message passing \cite{klicpera2020directional}, accounting for multi-scale structure in the network \cite{zhang2020molecular}, adoption of self-supervised learning \cite{wang2022molecular, schwaller2019molecular} and modeling interactions between geometric tensors \cite{batzner20223}.  

\begin{figure*}
  \includegraphics[width=0.98\textwidth]{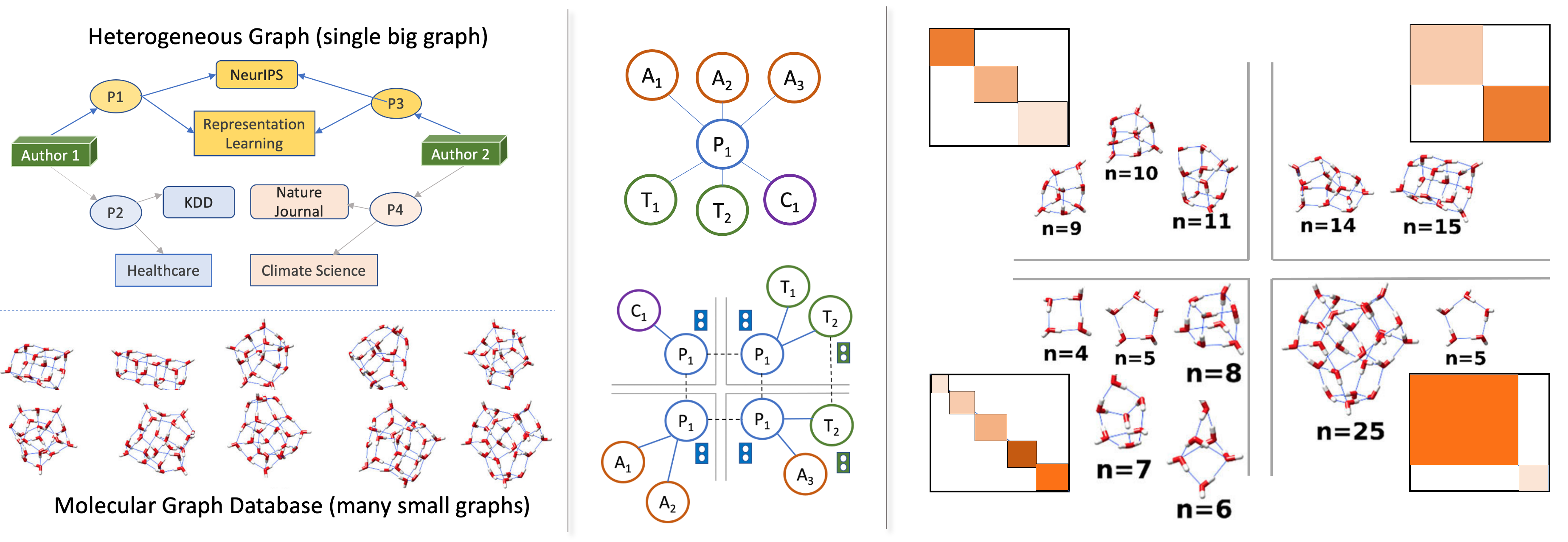}
  \vspace{-1em}
  \caption{Column 1: Top row shows a knowledge graph or "big graph"; bottom row shows a molecular database.  Column 2: Shows how big graphs are sampled (top row) and how nodes that falls across partition trigger communication during learning (bottom row).  Column 3: Shows how molecular graphs are packed into nearly 30-node packs.  Processing them requires support for efficient sparse block matrix operations.  Communications are quite localized.} 
  \label{fig:MolGNN-learning}
  \vspace{-1.5em}
 \end{figure*}
 
Fast training of these models becomes an increasingly important issue as the scientific community embraces these models to train on massive molecular databases \cite{schmidt2019recent, lim2019predicting, bilbrey2020jcp}. Evidence from the literature shows that these models are expensive to train, often spanning multiple days in multi-GPU or TPU settings \cite{bilbrey2020jcp, wang2022molecular, addanki2021large}. However, most of the research on scaling the training of GNN models have focused on learning over a single massive graph \cite{tripathy2020reducing, GNNAdvisor, kaler2022accelerating}, which subsequently steers the design space of software and hardware solutions \cite{zheng2020dgl, gandhi2021p3, zheng2022distributed} to addressing traditional graph processing challenges such as graph partitioning \cite{karypis1998multilevelk}, irregular memory access \cite{morari2014scaling} and workload imbalance due to power-law distributions \cite{geng2020awb}.  In contrast to training a GNN model on a single large graph (see Figure \ref{fig:MolGNN-learning}), molecular GNNs are trained on \textit{an ensemble of small individual graphs} with varying node counts. Molecular graph databases \cite{sussman1998protein, rakshit2019atlas, pinheiro2020machine, chanussot2021open} are characterized by the comparatively small size (both in terms of the total node count and the edge count) of individual graphs, and wide distribution of size and sparsity.  Additionally, most of these datasets represents graphs where each node is also associated with a 3D coordinate representing the position of an atom, which subsequently requires support for geometric operators such as nearest-neighbor computation.   

In a first work of its kind, we present a hardware-software co-design approach to accelerate the training of message passing graph neural networks that need to process millions of small graphs with diverse sizes.  Recent years have also witnessed the emergence of machine-learning accelerators such as tensor processing units \cite{jouppi2017datacenter}, reconfigurable dataflow  \cite{emani2021accelerating} and intelligence processing units (IPUs) \cite{jia2019dissecting} etc. The large on-chip high-bandwidth memory, support for fine-grained parallelism, and faster all-to-all communication links make the IPUs attractive for molecular GNN architectures, and is the primary target of this work.  

In addition to focusing on a specific architecture, we also identify a model to serve as a representative workload.  We focus on the SchNet neural network architecture ~\cite{SchNet2018, SchNetPack2019} for this paper. Given a molecular structure, SchNet constructs a nearest-neighbor graph around each atom within a radial cutoff distance (Figure \ref{fig:schnet}).  Next, it learns the representation of each atom by modeling pair-wise interactions between atoms based on a number of specified hops. Finally, a graph-level property such as the energy function is predicted by learning a composition function over each atom (or node) embedding.  SchNet and its variants have been extensively used for molecular property prediction in diverse domains such as computational chemistry \cite{bilbrey2020jcp}, drug discovery \cite{joshi20213d}, materials science \cite{jha2021enabling}.  Its pragmatic relevance and its capture of key kernels that are present in nearly all future model variants \cite{schwaller2019molecular, klicpera2020directional, batzner20223} make SchNet a compelling representative for molecular GNN-driven co-design.

Our paper makes the following contributions.
\begin{itemize}[noitemsep,topsep=0pt]
    \item We present a comprehensive study of a recently released benchmark dataset \cite{choudhury2020hydronet} that is among the largest 3D graph property prediction datasets available with highest degree of size and sparsity imbalance (\Cref{sec:char_mol_graph}).
    
    \item  We propose a hardware-agnostic technique namely \textit{batch packing} to coalesce small graphs while tracking their original connectivity structure for graph-level property prediction (\Cref{sec:batch_packing}). We show that \textit{batch packing} significantly reduces the memory footprint and improves the performance of the SchNet for diverse molecular databases with varying sizes and sparsity. 

    \item We implement a \textit{scatter-gather planner} for efficient scheduling of the gather and scatter operators of the molecular GNN on the IPUs (\Cref{sec:planner}). The planner finds a sweet spot between parallelism and device utilization. Additionally, we demonstrate the impact of other optimizations such as asynchronous, non-blocking I/O for batch loading, vectorization of scatter operation and pre-fetching for molecular GNN workloads (\Cref{sec:hw_optim}).
    \item  As a demonstration of extreme acceleration, we train the SchNet model on the Hydronet dataset \cite{choudhury2020hydronet} in 92 minutes using 64 IPUs, as compared to previously reported 2.7 days obtained 
    with a single-GPU setting \cite{bilbrey2020jcp}. Rapid training of scientific ML models is essential to accelerate scientific discovery, and we hope that our work will stimulate further research in this direction.
\end{itemize}

\section{Background: GNN-based Model for Quantum Chemistry}\label{sec:schnet_background}

\def\rcut{r_\text{cut}}
\def\edge#1#2{e_{#1#2}}
\def\vertex#1{v_{#1}}

\begin{figure}[t]
\centering
\vspace{-1ex}
  \includegraphics[width=0.4\textwidth]{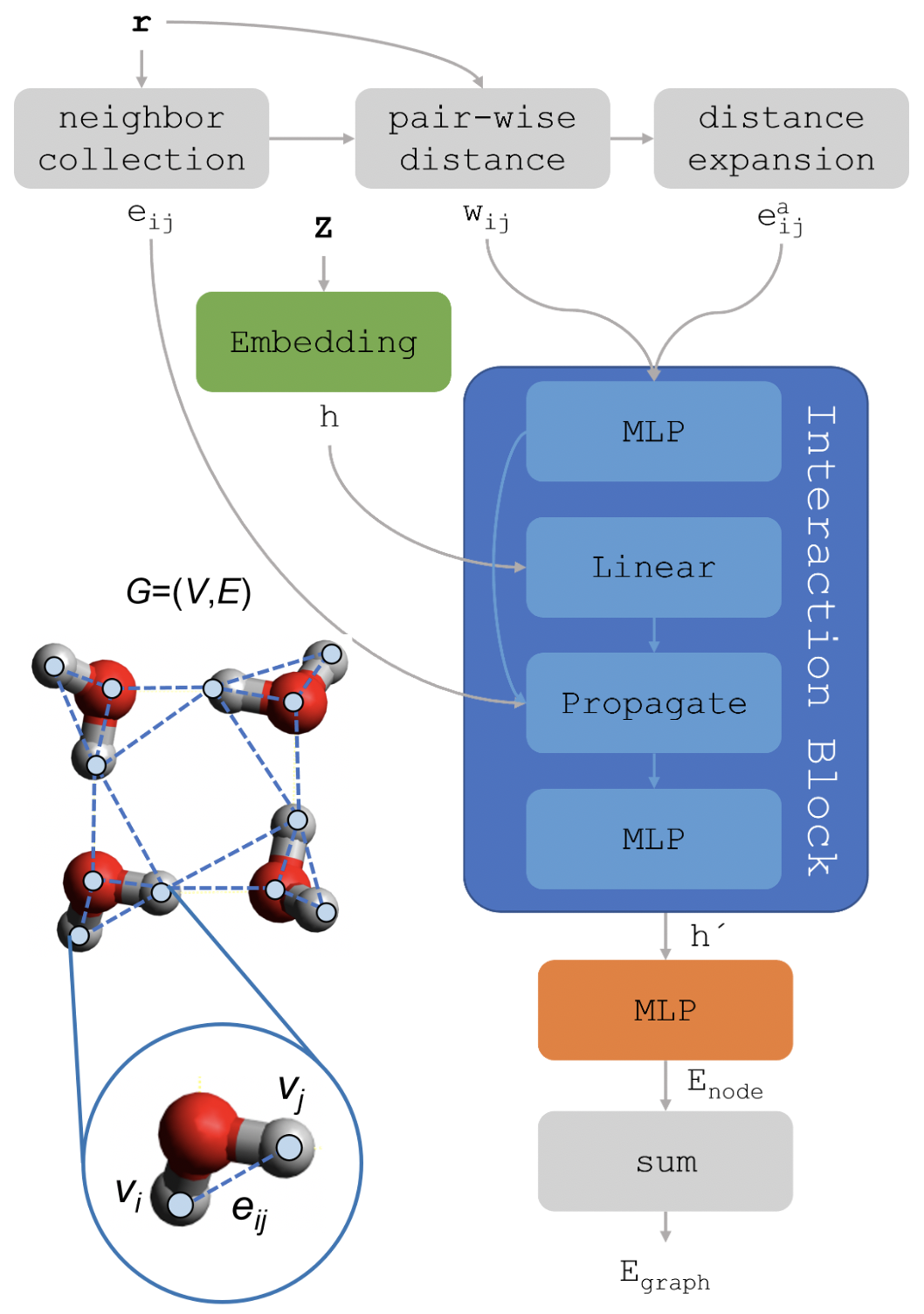}
  \vspace{-1em}
  \caption{Illustration of the GNN-based implementation of SchNet.}
  \label{fig:schnet}
  \vspace{-.5em} 
 \end{figure}

A molecule can be represented mathematically as a \textit{graph}, denoted by $G=(V,E)$, with vertex set~$V$ and edge set~$E$ (\Cref{fig:schnet}). Each vertex $\vertex i \in V$ represents an atom in the molecule and is associated with that atom's spatial coordinates~$\textbf r_i \in \mathbb R^3$, and atomic number~$z_i \in \mathbb N$.  The molecular properties calculated by \textit{ab initio} methods such as density functional theory (DFT) are determined entirely these node properties and rely on mean-field approximations to model complex many-body interactions \cite{KohnShamPhysRev.140.A1133}.  Explicit edges are introduced when applying GNNs to molecular property prediction tasks, where each edge~$\edge i j \in E$ reflects a pair-wise interaction between atoms~$\vertex i$ and~$\vertex j$.     The influence one atom has on another decreases non-linearly with the distance $d_{ij} = || \mathbf{r}_i - \mathbf{r}_j ||$ between the pair and beyond some critical distance $\rcut$ is known to be negligible.  In the context of electronic structure calculations this is often referred to as the \textit{nearsightedness of matter} \cite{Prodan_2005} where the specific value of $\rcut$ varies depending on the composition of the molecule or material.  We use this cutoff to define edges for the graph representation of a molecule as:
\begin{equation}
    \edge i j \in E \;\text{iff}\; d_{ij} < \rcut.
\end{equation}
In practice, a $K$-nearest neighbor (KNN) search is performed to return a fixed number of neighbors for each $v$.
As a consequence, the number of edges in the graph representation will grow at most linearly alongside the number of atoms in the molecule.


The SchNet GNN architecture \cite{SchNet2018} is trained to the learn mapping from this graph representation as $\text{SchNet}:\ G(V, E) \rightarrow \mathbb{R}$ where the prediction of the network is a property of the molecule.  The computation graph in this architecture has two phases (Figure \ref{fig:schnet}). The first phase focuses on atom-level representation learning, followed by an aggregation phase that learns a composition function of the set of atoms to predict a global property (such as the potential energy function). The entire computation process is visually described in \Cref{fig:schnet}.

\def\state{\textbf{h}}
\def\embedding{\mathsf{Embedding}}
\textsc{Embedding Layer:} Each vertex $\vertex i \in V$ represents an atom, with initial state~$\state_i = \embedding[z_i]$.


\def\ea{{e^\mathbf{a}}} 
\textsc{Interaction Block/GNN Layer:} The \textsl{interaction block} layers in Figure \ref{fig:schnet} are used to learn the atom-level representations via message-passing over the graph. The message to be passed combines states $\state$, and edge attributes $\ea_{ij}$, corresponding to a Gaussian radial basis function (RBF) expansion of $d_{ij}$:
\begin{equation}
\ea_{ij} = \Bigl[ \mathrm{exp} \bigl( -\gamma(d_{ij} - k\Delta \mu)^2 \bigr) \Bigr]_{k=0}^{N_\text{rbf}}
\end{equation}
where $\Delta \mu$ is the spacing of Gaussians with scale $\gamma$ on a grid ranging from 0 to $\rcut$, so $N_\text{rbf} = \rcut/\Delta \mu$.



The set of $\ea_{ij}$ are sent through a Multi-Layer Perceptron (MLP) to learn the `continuous filters', which are then weighted by a cosine function of $d_{ij}$ to reflect the nonlinear influence an atom has on its neighbor with respect to distance. Meanwhile, each node state $\state$ is passed through a linear layer, after which the states $\state$ and filters are propagated over edges $e_{ij}$ and passed through a subsequent MLP to obtain the updated node state~$\state'$: 
{
\setlength{\abovedisplayskip}{2pt}
\setlength{\belowdisplayskip}{0pt}
\begin{equation}
\label{equ:messagepassing}
\state_i' = \state_i + \sum_{j \neq i} f(\state_j, \ea_{ij})
\end{equation}
}

\textsc{MLP Layer:} The updated state $\state'$ is reduced through an MLP to a scalar value for each node, representing the contribution of each atom to the prediction target. 

\textsc{Activation and Pooling Layers:}
After the embeddings of all vertices in the final layer are computed, the contribution from all atoms are aggregated to obtain the global property of the molecular graph.

\section{IPU Hardware and Software}

The recently developed Graphcore IPU combines a number of hardware features with software abstractions to create a  platform that is exceptionally well suited to the unique computational demands of molecular GNNs.

\begin{figure}[!htbp]
   \vspace{-0.6em}
  \includegraphics[width=0.48\textwidth]{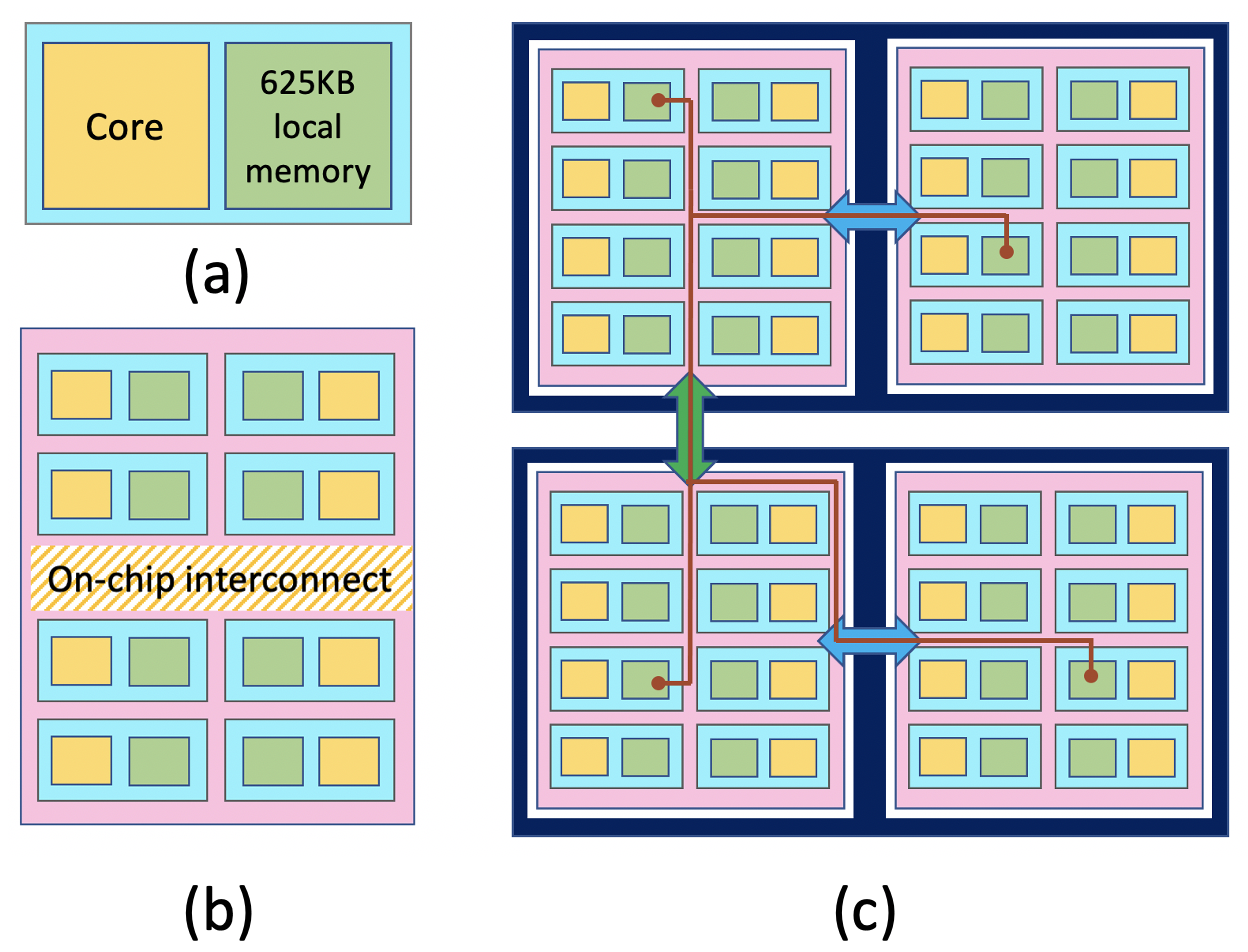}
  \vspace{-2em}
  \caption{An illustration of tile-based IPU processor architecture and communication topologies used for inter-card and inter-IPU communication.  Each card has two IPU processors. 
  } 
  \label{fig:ipu-memory-arch}
  \vspace{-1.2em}
 \end{figure}




\textbf{Hardware Architecture.}
The main components of a Graphcore IPU are the processing elements called \textit{tile}s. Each tile consists of one computing core and $\approx$ 625KB of local on-chip SRAM memory which is sufficient to store a large number of (packed) molecular graphs (Figure \ref{fig:ipu-memory-arch}a). The Bow IPU processor consists of a total of 1,472 tiles yielding a total of 900MB of high-bandwidth (65TB/s total) distributed SRAM memory available for machine learning workloads to exploit (Figure \ref{fig:ipu-memory-arch}b). Each computing core has 6 hardware threads that are round-robin multiplexed in time. 

An on-chip interconnect implements non-blocking, all-to-all pattern to enable high-bandwidth, low-latency communication among the tiles on an IPU. In addition, for inter-processor communication, low-latency, high-throughput \textit{IPU link}s are used (Figure \ref{fig:ipu-memory-arch}c). PCIe links are used to communicate with the host and in the Bow-2000 configuration, 4 IPUs are connected to a host CPU via PCIe connection (64GB/s) and form a IPU-POD system, and the communication between IPU are bypassing the high-throughput IPU-links (running at 320GB/s). 

The IPUs support bulk synchronous parallel (BSP) execution model at the hardware level. This model includes three phases: \textit{compute}: where each tile performs computation with the data available on its local SRAM memory, \textit{sync}: where threads are synchronized across all IPUs, and \textit{exchange}: where each thread exchanges data (located either on the same IPU but on different tile or on a tile of a remotely-located IPU).



\textbf{Support for ML Operators} The IPU is routinely programmed through high-level Python frameworks such as PyTorch and Tensorflow.  We use the PopTorch framework~\cite{Poptorch_git} which provides a set of extensions to PyTorch to efficiently target our implementation of SchNet to the IPU architecture.  We implement the interaction layer (section \ref{sec:schnet_background}) using the PyTorch-Geometric (PyG) framework \cite{fey2019fast}. An order-invariant gather function is used to aggregate the neighbor's influence on a given node via messages. The updated node embedding information is then \textit{scatter}ed back to the neighbors for the computation of the next set of embeddings in the next interaction layer. These gather-scatter operations for molecular graphs are localized on a single IPU so their execution is accelerated by utilizing the combined benefit of both the high-bandwidth local memory and the high-bandwidth communication over on-chip interconnect.  

The PopTorch framework performs compiler optimizations by leveraging both the structure of the model computation graph and the size of input tensors. After capturing an initial intermediate representation (IR), a number of compiler transformation passes are performed over this IR with the aim of  mapping high-level PyTorch modules to specific low-level instructions as well as the fusion of multiple constructs that can be performed more efficiently.  An example of an operator fusion pass that is essential for the efficient evaluation of the scatter-reduce operation commonly seen in message-passing GNNs is discussed in greater detail in \Cref{sec:vectorization_pass}.

Our adoption of the PyG framework is motivated by extensibility.  In principle, anyone should be able to extend our implementation of SchNet to more sophisticated models by altering the message-passing functions while still benefiting from the molecular GNN-targeted co-design features. 



\section{Co-Design Optimizations for Molecular GNNs}\label{sec:implementation_ipu}
Effective memory management for the data structures and intermediate computation, optimal parallelism and leveraging efficient all-to-all collectives are the key factors that determine the performance of molecular GNNs. We review a set of optimizations that we have implemented for the SchNet model that can be categorized into three broad categories: input-specific optimization (packing), hardware-targeted optimizations (planning, pre-fetching and compiler passes), and model-specific optimizations (merging weight updates and optimized softplus activation). In the following, we discuss these optimization techniques to improve the overall performance of the molecular GNNs.



\subsection{Input-Specific Optimization: Batch Packing}\label{sec:batch_packing}

Molecular graph structures vary significantly in  terms of the vertex counts and the edge counts. Efficiently targeting the high-bandwidth exchange fabric of the IPU is accomplished through static analysis of communication patterns between tiles for a given application.  This ahead-of-time compilation of molecular GNNs requires knowing the shapes of the input tensors as a priori for preparing the batches. To keep the shape sizes like the number of vertices and edges fixed,  
the naive strategy is to just pad batches up to the maximum number of vertices or edges (\Cref{fig:ipu_packing} (a)).
This can 
lead to a very large proportion of padding and therefore waste compute. 
GNN libraries such as PyG support the combination of multiple graphs into a single graph of multiple disconnected components.
However, without fixing the sizes of graphs within batches
no savings in padding can be made.
In this paper, we are the first to apply an approach, 
previously introduced in natural language
processing, where different sequences are concatenated~\cite{packing}.
However, instead of concatenating sequences/sentences, we combine graphs and 
instead of the number of words as the sequence length, we measure length
by the number of vertices,
as proposed in the original paper. We call this strategy \textit{batch packing}.
{
\begin{figure}[]
\vspace{-0.98em}
 \centering 
  \includegraphics[width=0.5\textwidth]{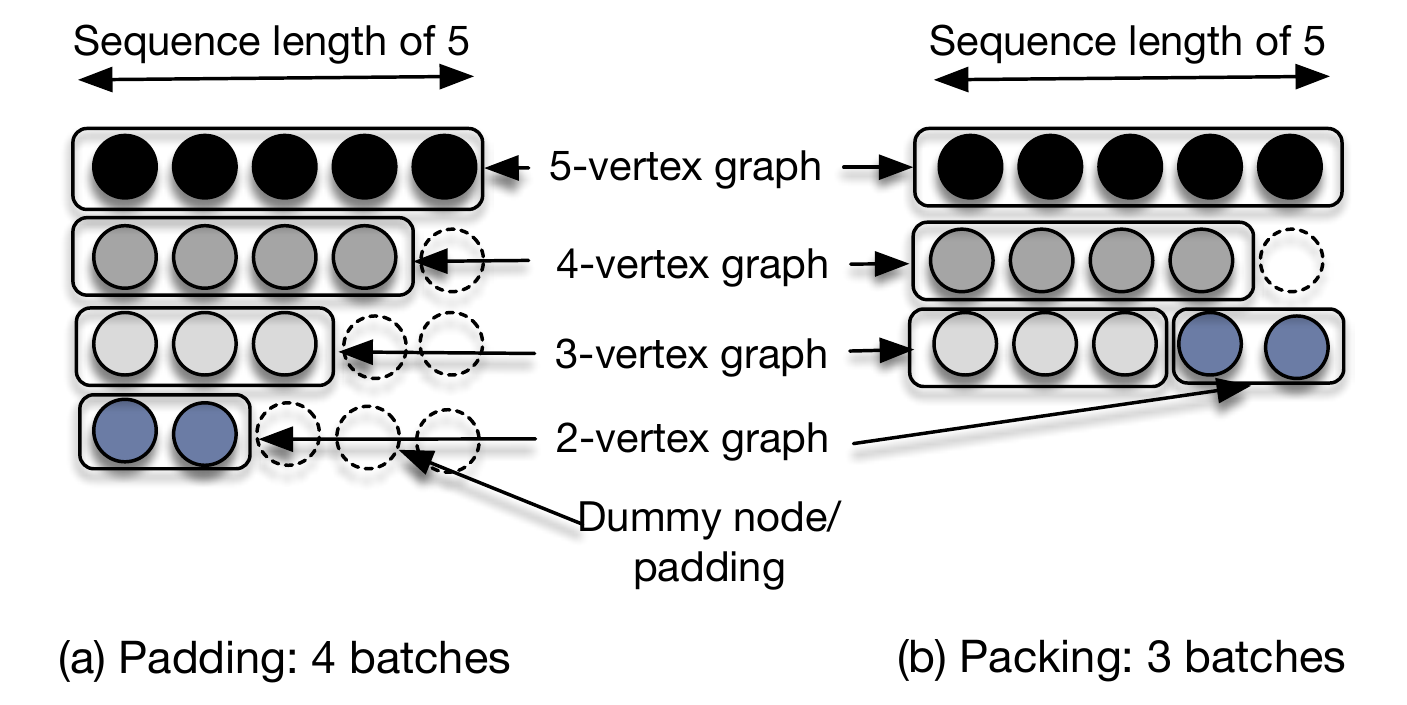}
\vspace{-2.5em}  
  \caption{Batch packing} \label{fig:ipu_packing}
\vspace{-1.3em}  
\end{figure}
}

Packing has been known to be 
NP-complete problem~\cite{Korte2012} for decades and is well established.
It can be formulated as
{
\setlength{\abovedisplayskip}{0.2pt}
\setlength{\belowdisplayskip}{0.2pt}
\begin{equation}
\begin{aligned}
\min_{y\in\{0,1\}^n,B\in\{0,1\}^{n\times n}} \quad & \sum_{j=1}^{n}{y_{j}} \\
\textrm{s.t.} \quad & \sum_{j=1} b_{ij} =1\quad   \forall i \\
  &\sum_{i=1}^n s(i)b_{ij}\leq s_m y_j \quad  \forall j.  
\end{aligned}
\end{equation}
}
We assume there are $n$ graphs to be packed 
and the combination of graphs is called \textit{pack}.
There can be a maximum of $n$ packs where $y_j\in\{0,1\}^n$ indicates 
which of the potential packs are actually used or not and is subject to minimization.
The assignment variable $b_{ij}\in\{0,1\}^{n\times n}$ indicates 
if the $i$-th graph gets put into the $j$-th pack.
The number of vertices in a graph is denoted by $s(i)$ and
$s_m$ is the maximum number of vertices in a pack.
This can be the maximum number of nodes in a graph or a larger number
as a trick to reduce padding.

{
\setlength{\textfloatsep}{0pt}
\setlength{\floatsep}{0pt}
\begin{algorithm}[htb]
\caption{Simplified LPFHP}\label{alg:LPFHP}
\begin{algorithmic}
\State \# Histogram counts
\State $h \gets \text{empty array}$
\For{$g_i \in Graphs$}
    \State $h[s(i)] \gets h[s(i)] + 1$
\EndFor
\State $S\gets\text{empty strategy dictionary of lists of pack counts}$
\For{$s \in [s_m, ..., 1]$}
    \State $c\gets h[s]$
    \State $R\gets s_m-s$
    \While{$c>0$}
        \State \# Get best fitting pack space and update strategy 
        \State $i = \min_{S[j]\neq\emptyset, j\in \{s, \ldots,s_m\}} {j}$
        \If{$i$ is NaN}
            \State $S[R] \gets [(c, [s])]$
            \State $c \gets 0$
        \Else
            \State $c, S \gets \text{update}(S, i, c, s)$
        \EndIf
    \EndWhile
\EndFor
\Return{S}
\State \textbf{def} update(S, i, c, s):
\State $c_p, p = S[i].pop()$
\If{$c\geq c_p$}
    \State $S[i-s].append([(c_p, p.append(s))])$
    \State $c \gets c-c_p$
\Else
    \State $S[i].append((c_p-c, p))$
    \State $S[i-s].append([(c, p.append(s))])$
    \State $c\gets 0$
\EndIf
\Return{c, S}
\end{algorithmic}
\end{algorithm}
\setlength{\textfloatsep}{0pt}
\setlength{\floatsep}{0pt}
}

In practice, numerous 
heuristic-based algorithm for packing
exist~\cite{johnson1973near,Lee1985,Johnson1985,Yue1995} that could be applied.
All these approaches are in the range of $O(n \log n)$ time  complexity,
where $n$ is the number of graphs/samples.
Hence, we use the faster approach from~\cite{packing}
that is called \textit{longest-pack-first histogram-packing} (LPFHP). 
The algorithm is derived from the best-fit method and is shown in~\Cref{alg:LPFHP}.
Best-fit means that an item gets added to the bucket that will leave the least space after the item was added.
The algorithm first computes the histogram  of graph sizes and respective sample counts. 
The algorithm also takes as input the potential maximum sequence lengths in a pack, $s_m$. 
Next, it iterates from the largest to the smallest graph and tries to add it to existing combined graphs/packs.
The trick here is that the algorithm works on histogram bins and counts and not single graphs.
If a graph can fit with multiple different already combined graphs, we chose the one with the least space (best-fit). 
For example, if we need to add a graph with size 10 and our maximum is 100, 
we would prefer to combine it with a graph of 90 nodes to get a perfect match instead of combining it with a pack or graph of size 11, 
which is the smallest setting that can occur. 


\vspace{-0.5em}
\subsection{Hardware-targeted Optimizations}\label{sec:hw_optim}
For accelerating the training of molecular GNN models on the IPUs, we leverage fully static compilation and a gather-scatter planner (\Cref{sec:planner}). With static compilation, access to the computation graph provides the opportunity to reduce maximum memory usage by changing the order of operations (called ``rescheduling'') in a full compute graph to reduce the maximum live memory at any given point in the program. This both may allow a full compute graph to fit into SRAM, and increase the amount of available memory at a given point in the program. Also, increased SRAM availability can allow some operations to make better use of high bandwidth local memory to accelerate operations and assist the planner to decide optimal data layout. At the compiler level, two IPU-specific features are implemented for optimizing the execution of the molecular GNNs: scatter vectorization, and the planning of scatter/gather operators for targeting the IPU.

\subsubsection{Vectorization of gather/scatter operations}\label{sec:vectorization_pass}

One of the benefits of the ahead-of-time compilation of a PyTorch model is the ability to write specialized passes that can perform operator fusion to efficiently target the IPU hardware.  These passes can apply a simple pattern matching on the captured computation graph and replace or fuse multiple operations into new patterns that are known to be more efficient to evaluate.  In the context of GNNs, we have developed a fusion pass that finds a sequence where a vector of indices is broadcasted ahead of aggregation.  Our pass removes the redundant broadcasting step and ensures that the more efficient vectorized form of the scatter operation is targeted by the scatter/gather planner.  

\subsubsection{The Scatter/Gather Planner}\label{sec:planner}
The message passing layer that forms part of the Interaction Block/GNN layer discussed in~\Cref{sec:schnet_background} \textit{gather}s the embeddings for atoms in the neighborhood $N(v)$ of each atom, $\state_j$, as well as the generated `continuous filter' values corresponding to each edge in the graph, $e_{ij}$, for the aggregation phase of the GNNs.


Gathering yields the values of each row in a matrix $A$ corresponding to each row index in a vector $i$ yielding a matrix $\text{gather}(A, i)$.
{
\setlength{\abovedisplayskip}{0.1pt}
\setlength{\belowdisplayskip}{0.1pt}
\begin{equation}
\begin{split}
\label{equ:gather}
A \in &\ \mathbb{R}^{M{\times}N} \\
i \in &\ \mathbb{N}^I \\
\text{gather}(A,i) \in &\ \mathbb{R}^{I{\times}N} \\
\text{gather}(A,i) =& \ \begin{bmatrix}
  A_{i_0,0} & A_{i_0,1} & \cdots & A_{i_0,N} \\
  A_{i_1,0} & A_{i_1,1} & \cdots & A_{i_1,N} \\
  \vdots & \vdots & \ddots & \vdots \\
  A_{i_I,0} & A_{i_I,1} & \cdots & A_{i_I,N}
\end{bmatrix} \\
\end{split}
\end{equation}
}

Once messages have been calculated they are \textit{scatter}ed so that messages for each destination atom $v$ are aggregated together which gives the output embeddings $h'$ for the layer shown in Equation \ref{equ:messagepassing}.

Scattering takes as input a matrix $A$ and a set $U$ of sparse matrices which are all aggregated. The sparse matrices $U$ are collectively represented as a list of non-zero row indices $i$ and a list of non-zero row values $V$.
{
\setlength{\abovedisplayskip}{0.1pt}
\setlength{\belowdisplayskip}{0.1pt}
\begin{equation}
\begin{split}
\label{equ:scatter}
A \in &\ \mathbb{R}^{M{\times}N} \\
i \in &\ \mathbb{N}^I \\
V \in &\ \mathbb{R}^{I{\times}N} \\
\text{scatter}(A, U) \in &\ \mathbb{R}^{M{\times}N} \\
\text{scatter}(A, U) =&\ A + {\sum_{B \in U} B}
\end{split}
\end{equation}
}

\textit{Gather} provides batch read and \textit{scatter} provides batch read-modify-write of dynamically indexed chunks of memory. These operations' runtimes are dominated by memory access.
Optimised implementations of \textit{gather} and \textit{scatter} for IPU hardware are used to coordinate dynamically indexed memory access across the distributed tile memories and maximise memory bandwidth utilization.

\textbf{Planning.} The IPU scatter/gather implementation is controlled by the scatter/gather planner. This is a host utility that minimizes a cost function for a scatter/gather operation by varying implementation parameters. Parameters of the implementation that may be varied control how an individual scatter or gather operation is parallelized across tiles in the IPU. 


This implementation of both gather and scatter uses a divide and conquer strategy. The full gather or scatter operations are divided evenly into smaller gather or scatter operations that are locally executed on each tile. The results are then exchanged and optionally reduced to give the final result. The terms $P_I$, $P_M$, and $P_N$ represent divisors for 3 dimensions of the full gather or scatter operation $I$, $M$, and $N$ outlined in Equations \ref{equ:gather} and \ref{equ:scatter}. $I_t = \left \lceil{\frac{I}{P_I}}\right \rceil$, $M_t = \left \lceil{\frac{M}{P_M}}\right \rceil$, and $N_t = \left \lceil{\frac{N}{P_N}}\right \rceil$ give the size of the sub-problem, or partition, calculated on each tile:
{
\setlength{\abovedisplayskip}{0.1pt}
\setlength{\belowdisplayskip}{0.1pt}
\begin{equation}
\begin{split}
A_t \in &\ \mathbb{R}^{M_t{\times}N_t} \\
i_t \in &\ \mathbb{N}^{I_t} \\
V_t \in &\ \mathbb{R}^{I_t{\times}N_t} \\
\text{gather}(A_t, i_t) \in &\ \mathbb{R}^{I_t{\times}N_t} \\
\text{scatter}(A_t, i_t, V_t) \in&\ \mathbb{R}^{M_t{\times}N_t}
\end{split}
\end{equation}
}

\textbf{Cost Function for Planner}
The cost function used for planning (\Cref{sec:planner}) is one that estimates maximum machine cycles required to complete the operation on any one tile and a minimum is found by exhaustive search of valid implementation parameter settings. Simplified equations for gather and scatter planner cost $c_\text{gather}$ and $c_\text{scatter}$ are listed in Equations \ref{equ:plannergathercost} and \ref{equ:plannerscattercost}.
{
\setlength{\abovedisplayskip}{0.1pt}
\setlength{\belowdisplayskip}{0.1pt}
\begin{equation}
\begin{split}
\label{equ:plannergathercost}
B_\text{data} =\ & \text{bytes per data element} \\
B_\text{index} =\ & \text{bytes per index element} \\
B_\text{vwidth} =\ & \text{tile load/store/acc bytes per cycle} \\
W =\ & \text{number of worker threads per tile} \\
I_t =\ & \left \lceil{\frac{I}{P_I}}\right \rceil \\
M_t =\ & \left \lceil{\frac{M}{P_M}}\right \rceil \\
N_t =\ & \left \lceil{\frac{N}{P_N}}\right \rceil \\
e(b) =\ & \frac{b}{\text{tile exchange send/receive bytes per cycle}} \\
g(i, m, n) =\ & W\left \lceil{\frac{i}{W}}\right \rceil \frac{nmB_\text{data}}{MB_\text{vwidth}} \\
c_\text{partial gather} =\  & e(M_tN_tB_\text{data}) + e(I_tB_\text{index}) + \\
                    & g(I_t, M_t, N_t) \\
c_\text{gather reduce} =\ & e(I_tN_tB_\text{data}) + \frac{I_tN_tB_\text{data}}{B_\text{vwidth}} \\
c_\text{gather} =\ & c_\text{partial gather} + \begin{cases}
   c_\text{gather reduce}, & \text{if}\ P_M > 1 \\
   0, & \text{otherwise}
\end{cases}
\end{split}
\end{equation}

\begin{equation}
\begin{split}
\label{equ:plannerscattercost}
s(i, m, n) =\ & W\left \lceil{\frac{m}{W}}\right \rceil \frac{inB_\text{data}}{MB_\text{vwidth}} \\
c_\text{partial scatter} =\ & e(I_tN_tB_\text{data}) + \\
                            & e(I_tB_\text{index}) + \\
                            & u(I_t, M_t, N_t) \\
c_\text{scatter reduce} =\ & e (M_tN_tB_\text{data}) + \frac{M_tN_tB_\text{data}}{B_\text{vwidth}} \\
c_\text{scatter} =\ & c_\text{partial scatter} + \begin{cases}
  c_\text{scatter reduce}, & \text{if}\ P_I > 0 \\
  0, & \text{otherwise}
\end{cases}
\end{split}
\end{equation}
}

Here $g$ represents a function estimating machine cycles taken to execute a gather on a single tile in terms of the SRAM load/store bandwidth $B_\text{vwidth}$. $s$ represents a function estimating machine cycles taken to execute a scatter on a single tile in terms of the SRAM load/store bandwidth and accumulation vector width $B_\text{vwidth}$. Note that these equations omit many overheads in the implementation for simplicity and represent more of a theoretical minimum. The real cost function used by the planner accounts for more of these overheads and practical implementation details. The function $e$ simply represents how many cycles a certain number of bytes would take to be sent or received on a tile.


There are different tradeoffs when varying $P_I$, $P_M$, and $P_N$ captured by this cost function. Increasing one of these may increase the amount of data that must be communicated between tiles. Increasing another may require a more costly reduction after computing per-tile partition results. Note that when all data, indices, and results fit into SRAM and we can use this implementation of gather or scatter, data never need be read from/written off-chip. Inter-tile communication under this implementation is always balanced allowing every tile to fully utilise its receive bandwidth thanks to regular communication patterns between tiles, and balanced sent/received payloads on each tile. Under the assumption that row indices are i.i.d. along the indexable dimension $M$ load/store/accumulate hardware is fully utilised during on-tile gather, scatter, or reduction steps.

For both gather and scatter operations, this results in an initial exchange of data between tiles to provide the input to each tile's partition, followed by on-tile computation to calculate the result for each tile's partition of the full problem.
A reduction of the results of either $\text{gather}(A_t, i_t)$ or $\text{scatter}(A_t, i_t, V_t)$ for each tile will then require further reduction to get the final result when $P_M > 1$.  This is necessary as the full range of the indexed dimension $M$ is not available on each tile in this case. The additional reduction entails an all-to-all exchange of data between tiles to redistribute partial results (there will be $P_M$ partial results spread over of $P_M$ tiles), followed by on-tile computation to reduce the $P_M$ partials to get the final result.

\vspace{-0.5em}
\subsubsection{Host-Device I/O Optimizations}
\textbf{Asynchronous, non-blocking I/O} The preparation of mini-batches can be expensive as it involves the random access of irregular sized molecular graphs followed by the collation process to combine multiple graphs into one large disconnected graph.  We can improve the time to access a single graph by using a two-level caching strategy:
\begin{itemize}[noitemsep,topsep=0pt]
    \item molecular graphs are stored on disk in an efficient compressed serialized binary representation for multidimensional tensor data.
    \item the fully materialized graph data structure is cached in memory on first time access which helps reduce redundant disk I/O.
\end{itemize}
We apply this caching strategy within multiple asynchronous workers for preparing mini-batches on the host, which effectively overlaps the time for preparing mini-batches with the computation on the IPU.

\textbf{Pre-fetching} To reduce the waiting time between host-device transfers, a \textit{pre-fetch depth} can be specified for the data stream transfer. The depth dictates the number of pre-fetched batches and enables the host-to-device transfer of the next mini-batch simultaneously while the device continues evaluation of the training loop on the current mini-batch.

\subsection{Model-specific Optimizations}\label{sec:model_specific_optim}

\textbf{Merged Communication Collectives.} During data parallel training, each replica (IPU) computes a local gradient for the mini-batch.  
These gradients are then combined across all replicas ahead of performing the weight update. 
To reduce the latency in performing the weight update, all reduce collectives are merged together to communicate all variables at once.
The effect of this optimization can be observed from the profiler output included in~\Cref{fig:merge_allreduce_profile} in the Appendix. 


\textbf{Optimized softplus.}
The default implementation of the softplus activation is defined in PyTorch as:
{
\setlength{\abovedisplayskip}{0.1pt}
\setlength{\belowdisplayskip}{0.1pt}
\begin{equation}
    \label{equ:torch_softplus}
    \text{softplus}(x) = 
    \begin{cases}
        \frac{1}{\beta} \log(1 + \exp(\beta x)), & \text{if } \beta x \leq \tau \\
        x, &\text{if } \beta x > \tau,
    \end{cases}
\end{equation}
}
where $\tau$ is a threshold value for which the activation saturates to the linear function.  This conditional expression is used for numerical stability but for the default values of $\tau = 20$ and $\beta = 1$ we can more efficiently evaluate the stable softplus activation equivalently as follows:
{
\begin{equation}
    \text{softplus}(x) = \log(1 + \exp(-|x|)) + \max(x, 0),
\end{equation}
}
which is used in our optimized implementation of the SchNet model as this simplifies expression compiles down to a more efficient compute vertex than the original formulation in Equation \ref{equ:torch_softplus} and is numerically stable without additional parameters.

\begin{figure}
\centering
  \includegraphics[width=0.35\textwidth]{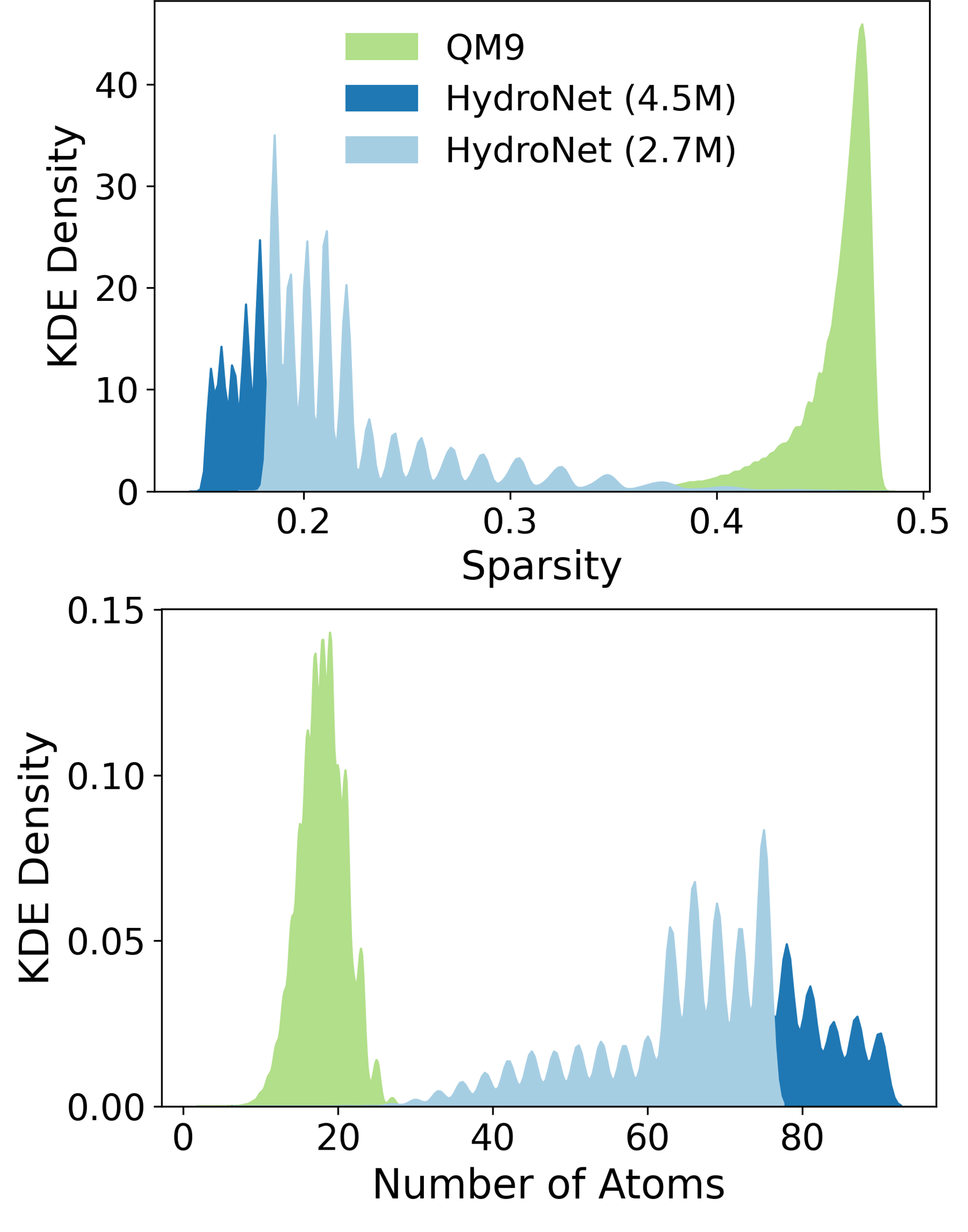}
  \vspace{-1.2em}
  \caption{Characteristics of the HydroNet~\cite{HydroNetDataset} and QM9~\cite{ramakrishnan2014quantum} benchmark datasets and example molecular graph $G$ for a water cluster with 12 vertices. Here KDE stands for kernel density estimate. Smaller sparsity value implies sparser graph.}\label{fig:sparsity}
\vspace{-0.5em}  
 \end{figure}

\vspace{-1.2em}
\section{Evaluation}
We performed a comprehensive set of experiments to evaluate the effect of our proposed optimizations for the SchNet model to predict the molecular properties on a large-scale IPU system. In addition, we conducted strong scaling experiments and benchmarked our holistic co-designed approach with an out-of-the-box SchNet implementation modified to run on multiple GPUs. These results are discussed in the following subsections.   
\vspace{-0.5em} 
\subsection{Experimental Setup} 
\subsubsection{Hardware Configuration}
The GPU experiments were conducted on Ampere A100 GPUs. This system consists of 8 A100 GPUs with 
40 GB
main memory attached to each GPU.
All IPU experiments were conducted on a Bow Pod64 IPU system.  We used this system to study the scaling behavior for data parallel training from a single IPU up to 64 IPUs.
\vspace{-0.5em} 
\subsubsection{Hyperparameter setup}
We used an up-to-date SchNet model implementation in PyTorch Geometric 
for all experiments reported here.  Except where otherwise noted, we used 4 interaction blocks, a hidden feature size of 100, and a uniform grid of 25 Gaussians for the basis function expansion of the inter-atomic distances.  This model was compiled for IPUs using PopTorch version 3.0.0 which uses PyTorch version 1.10.1 implicitly.  We used the Adam optimizer with a learning rate of 0.001 and collected throughput and speedup numbers by executing twenty-five epochs of a standard training loop.

\vspace{-0.8em} 
\subsection{Characterization of the Molecular Graphs}\label{sec:char_mol_graph}
In this work, we demonstrate the scalability of the accelerated SchNet model on the following two datasets with diverse characteristics.

\textsc{HydroNet:} The Molecular Property Prediction task outlined in the HydroNet benchmark \cite{HydroNetDataset} is as follows: given a molecule with specified spatial coordinate information, predict its energy. The HydroNet benchmark dataset is made up of 4.5M water cluster samples/molecules. Each sample molecule (graph) has between 9 and 90 atoms (vertices) and the task is to predict the energy \textit{E} of the cluster. 
Of particular interest is the sparsity of the water cluster graphs, as demonstrated in 
\Cref{fig:sparsity}. Because edges are applied between nodes based on a limited spatial distance, physical constraints limit the number of atoms that can be packed into a region of space. Therefore, as the size of the cluster increases, so does the sparsity of the graph. We examine the full benchmark dataset, as well as a 2.7M subset containing clusters of size 9--75 nodes to examine the effect of reduced sparsity.

\textsc{QM9:} In contrast, the widely used QM9 molecular database \cite{ramakrishnan2014quantum}, which is limited to molecules with a maximum of 29 atoms, shows less sparsity in the corresponding molecular graphs (\Cref{fig:sparsity}). The limited number of atoms is the cause of the relatively dense molecular graph; however, increasing the number of atoms to examine, for example, liquid or condensed solid phases, would necessarily lead to increased sparsity similar to that of the HydroNet dataset. Because such small molecule datasets are commonly applied to train molecular NNs, we compare against the performance on the QM9 dataset. 

These benchmarks are broadly representative for the current generation of electronic structure methods that are used to generate the training data for Molecular GNNs.  One of the most widely used approximations in \textit{ab initio} methods is density functional theory (DFT) which when applied to the modelling of molecules and materials is limited to systems of approximately one hundred atoms in routine calculations.  This limit is not expected to be overcome any time soon despite decades of progress on linear scaling DFT methods that have yet to deliver the promise of quantum mechanical accuracy for molecules and materials composed of several thousands of atoms \cite{KohnPhysRevLett.76.3168, prentice_onetep_2020}.

\begin{figure}
    \centering
    \includegraphics[width=0.44\textwidth]{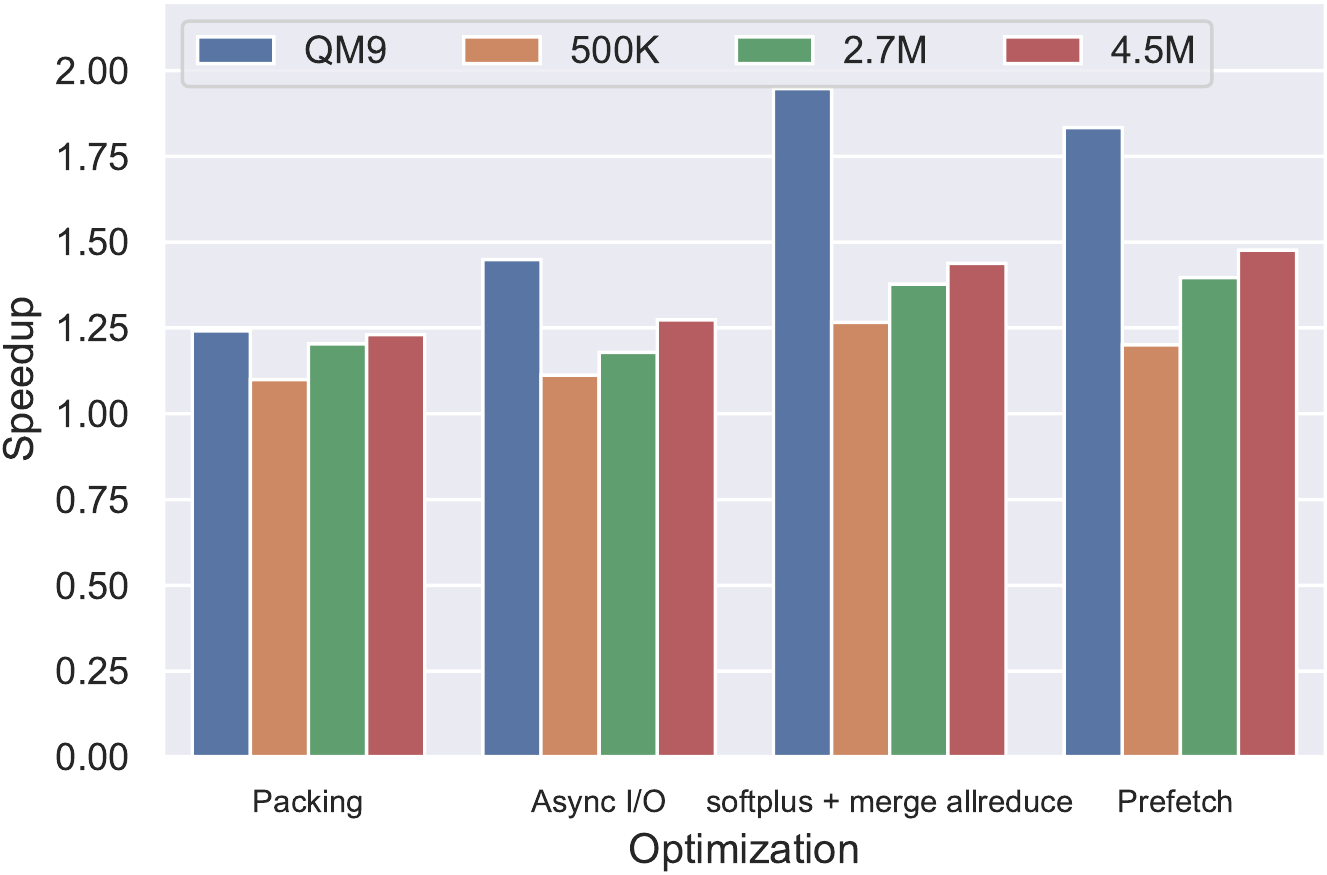} 
\vspace{-1.3em}    
    \caption{Speedup of the IPU SchNet model with different optimizations  w.r.t. a baseline IPU implementation. The optimizations are added progressively from left to right. For example, the legend Async/IO implies packing with Async/IO. The experiments were conducted on 16 IPUs.}
    \label{fig:speedup_perf_ablation}
\vspace{-1.5em}
\end{figure}

\begin{figure}
  \centering
    \begin{subfigure}[t]{0.23\textwidth}
  \centering
  \includegraphics[width=\textwidth]{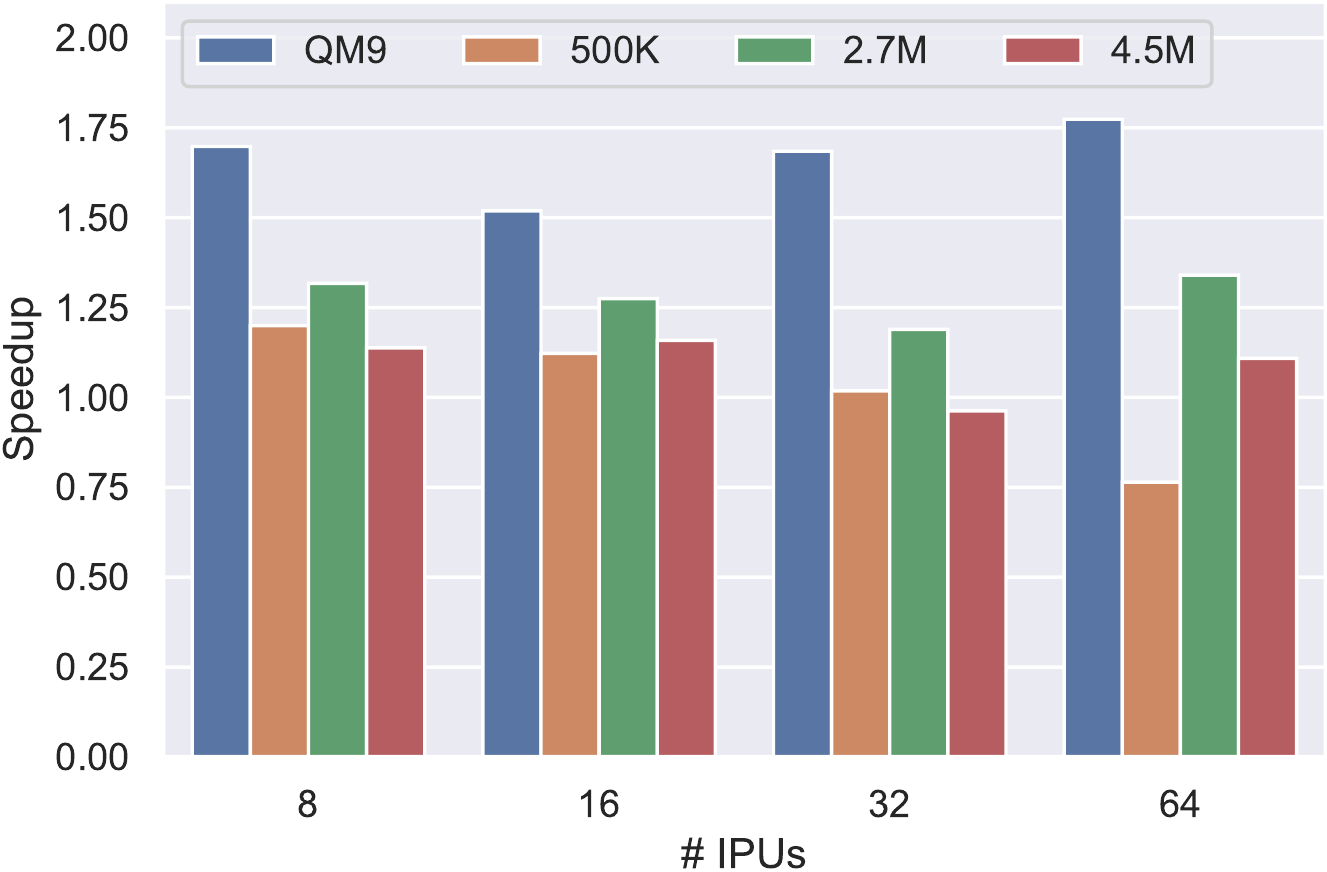} 
    \caption{ Packing over padding.}\label{fig:packing_padding_scale}
\end{subfigure} %
  \begin{subfigure}[t]{0.23\textwidth}
  \centering
    \includegraphics[width=\textwidth]{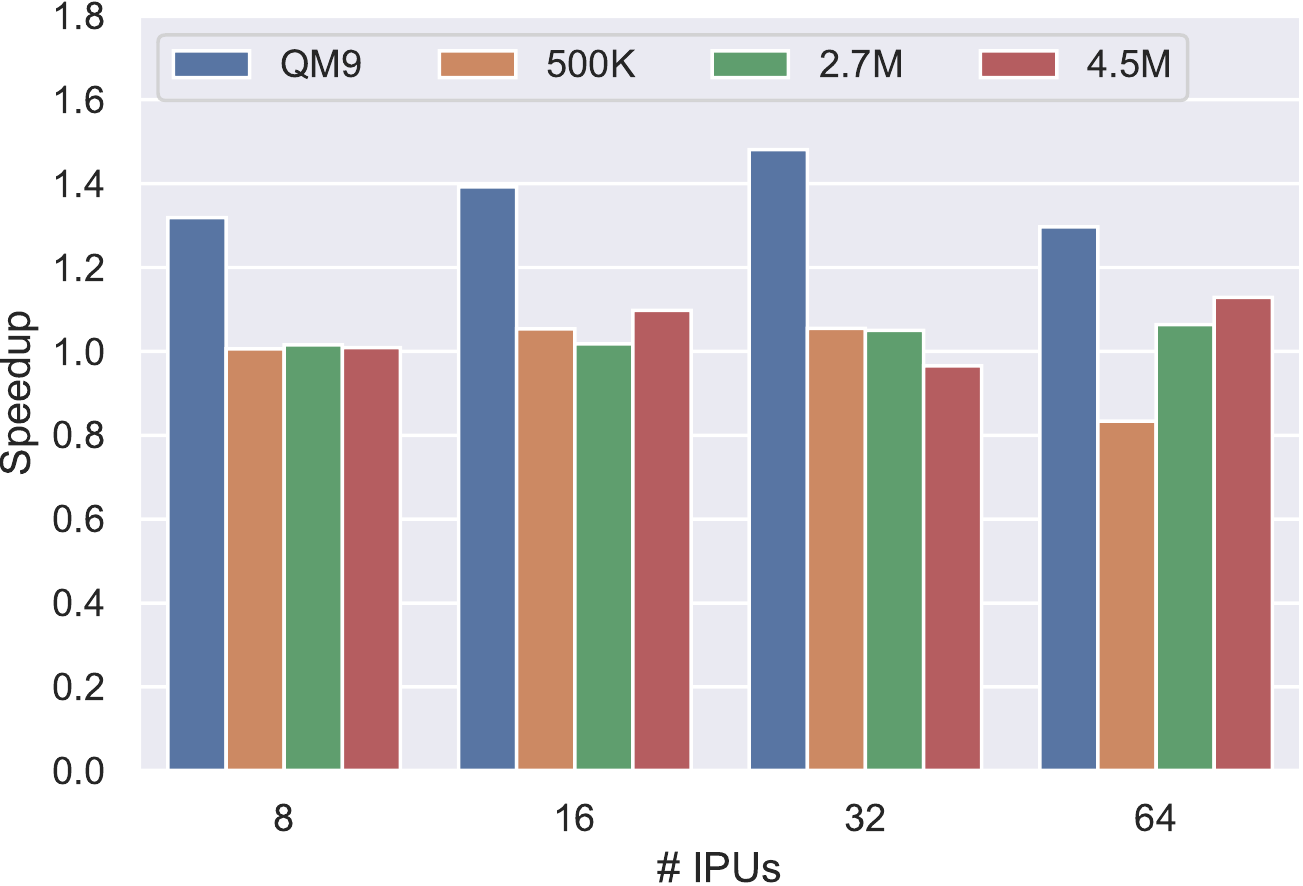} 
    \caption{Asynchronous I/O over the synchronous dataloader.}
    \label{fig:speedup_async_dataloader_scale}
 \end{subfigure}
 \vspace{-1em}
 \caption{Effect of packing and asynchronous I/O at different scales. Here, all additional optimizations are enabled.}
\end{figure}

\vspace{-1.1em} 
\subsection{Evaluation of Different Optimization Techniques}
\subsubsection{Evaluation of the Packing Technique}
We first evaluate the performance of the IPU SchNet model with both packing and padding. The experimental results are reported in~\Cref{fig:speedup_perf_ablation} and in~\Cref{fig:packing_padding_scale}. The speedup is calculated by taking the ratio of the average model training time with padding and packing. As can be observed from~\Cref{fig:speedup_perf_ablation}, packing may improve the performance of the SchNet model by upto ~25\%, compared to the original padding. As the number of IPUs are increased (~\Cref{fig:packing_padding_scale}), packing technique becomes more impactful with the QM9 dataset due to its denser structure compared to the water cluster graphs as well as  lower node counts (\Cref{fig:sparsity}). With larger 4.5M water cluster dataset and 64 IPUs, the packing technique outperforms the model with padding, as the compaction of the graphs enables processing more batches concurrently with more compute power.

\begin{figure}
    \centering
    \includegraphics[width=0.45\textwidth]{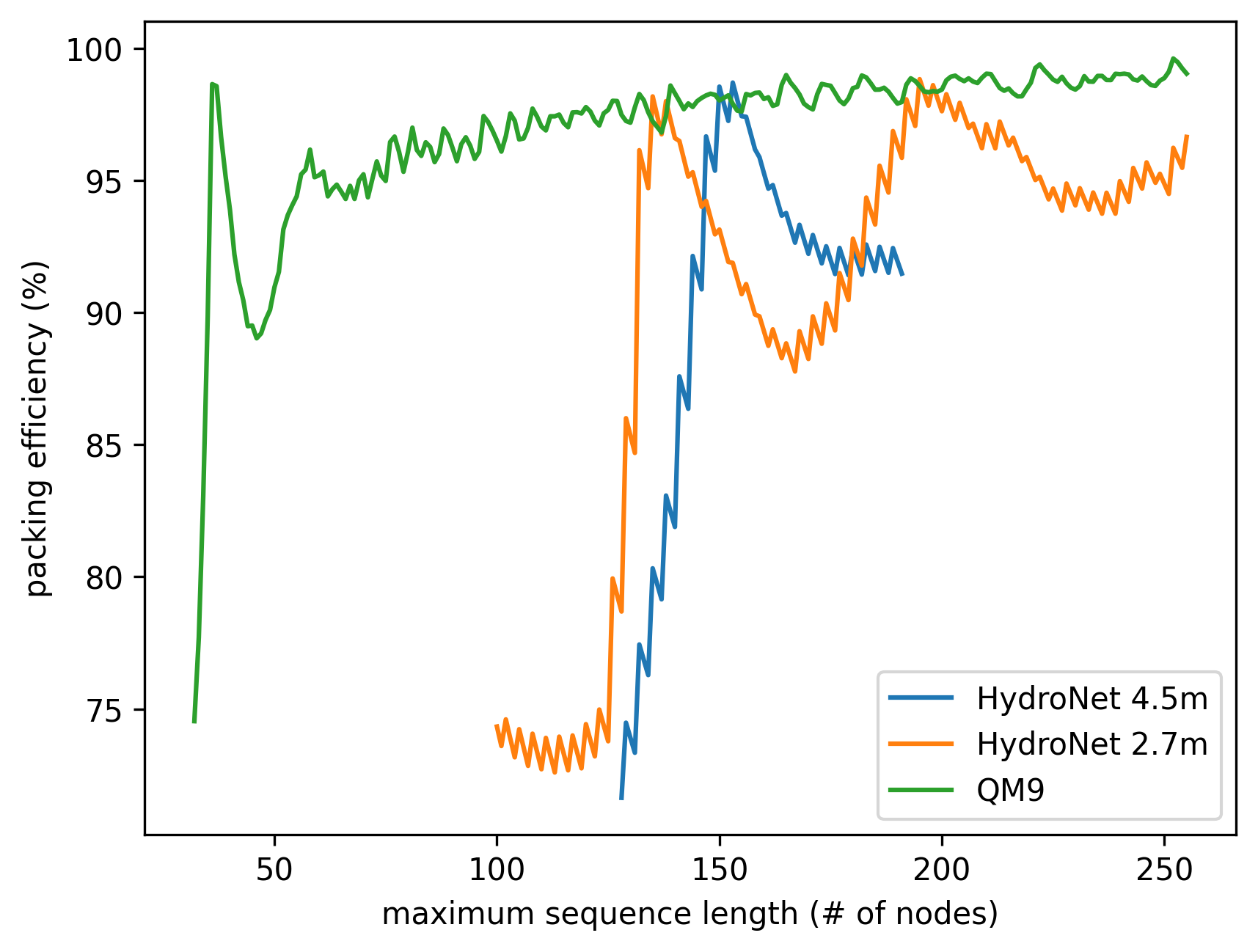} 
    \vspace{-1.5em}
    \caption{Effect of the maximum number of vertices in a pack (sequence length) on packing efficiency.}
    \label{fig:qm9schnetpack}
 \vspace{-1.0em}
\end{figure}

We also evaluate the efficiency of our LPFHP batch packing algorithm compared to the padding technique (\Cref{fig:qm9schnetpack}). Efficiency is defined as the percentage of padding reduced by applying the LPFHP batch packing algorithm. With the QM9 dataset, padding may result in $38\%$ wastage of memory  where the maximum number of vertices in a graph is 29 (\Cref{fig:qm9schnetpack,fig:sparsity}). To determine the optimal maximum sequence length for packing, it is to be noted from the histogram of the number of nodes in graphs that  
sometimes the mode of the distribution 
is larger than half of the maximum number of nodes, as is the case for the QM9 as well as HydroNet dataset (\Cref{fig:sparsity}).
In this case, if the maximum sequence length, i.e. maximum number of nodes in a pack, $s_m$ is set to a smaller value, a substantial amount of padding would still 
be required ($30\%$ on QM9 compared to $38\%$ with normal padding). 
Thus, it is beneficial to increase the allowed maximum number of nodes
in a pack of graphs that gets processed jointly.
For QM9, this can reduce the padding to less than $2\%$ (\Cref{fig:qm9schnetpack}).
Similar observation can be made for the two HydroNet datasets.
In all cases, the curves are not smooth and contain several systematic spikes.
This can be explained with the histograms, displayed in ~\Cref{fig:sparsity}.
Only certain numbers of nodes occur in the dataset and there are several gaps 
and even for the existing numbers, the distributions are not smooth.
This directly translates into the packing results.

\subsubsection{Impact of Asynchronous I/O}
In addition to applying packing technique, enabling asynchronous I/O improves the performance of the SchNet model, as reported in~\Cref{fig:speedup_perf_ablation} and \Cref{fig:speedup_async_dataloader_scale}. We observe that, with these two techniques, the QM9 dataset achieves better improvement in performance.

\subsubsection{Impact of the optimized softplus, merging allreduce and prefetching}
Adding a cycle optimized implementation of the softplus activation and merging the all-reduce communincation collective improves the performance of training the SchNet model using IPUs for all the datasets (\Cref{fig:speedup_perf_ablation}). While pre-fetching improves performance with the 4.5M water cluster dataset, it negatively impacts the performance of the QM dataset. Here, the prefetch depth is set to 4.

\begin{figure}
  \includegraphics[width=0.5\textwidth]{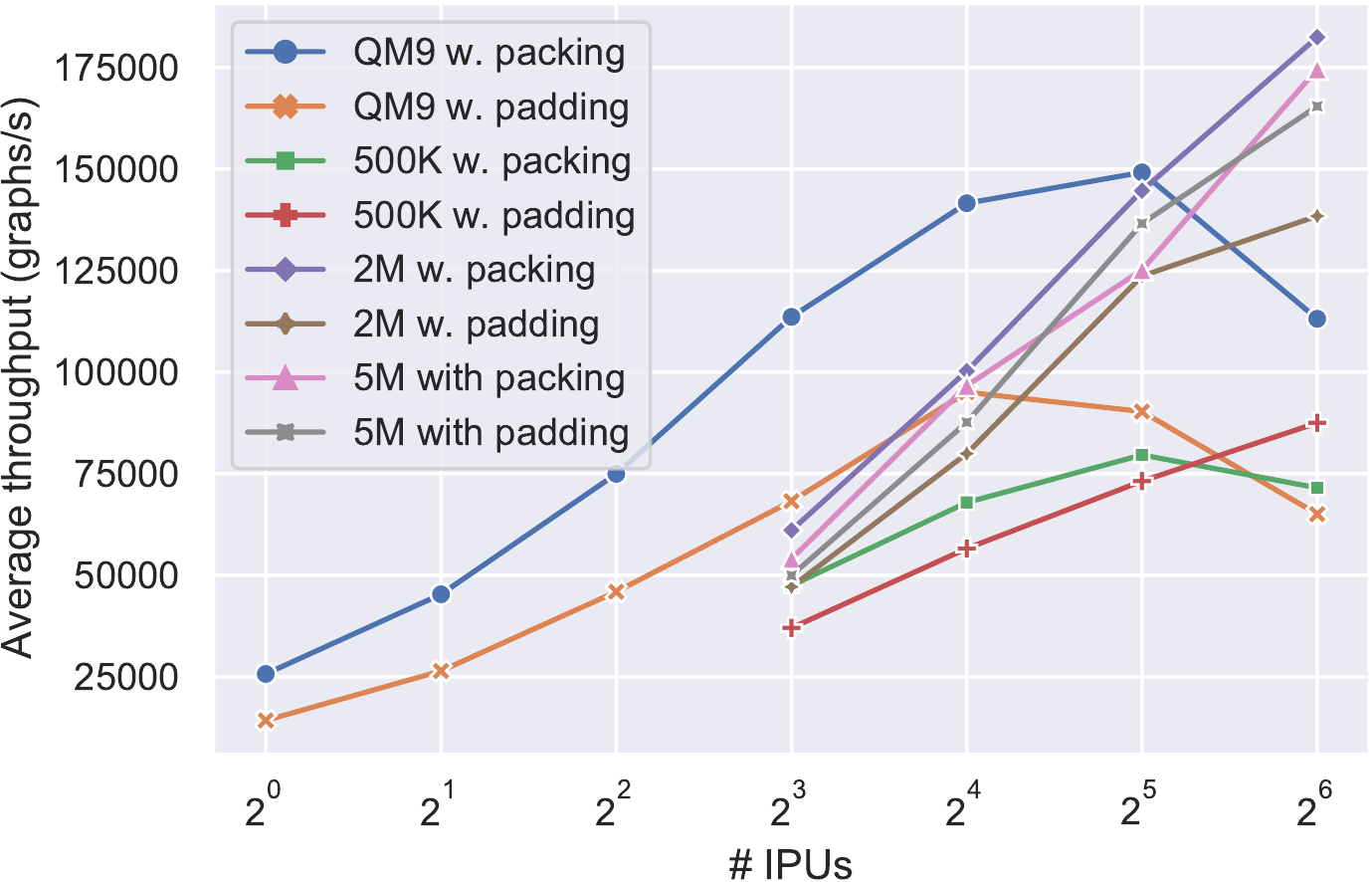}
\vspace{-1.6em}  
  \caption{Strong scaling results of the optimized SchNet model on the IPUs with packing and padding. All additional optimizations are also applied in both cases. 
  } \label{fig:schnet_strong_scaling}
\vspace{-1.2em}  
\end{figure}

\subsection{Strong Scaling Results}
We report the strong scaling training performance of the SchNet model on the IPUs in~\Cref{fig:schnet_strong_scaling}. Here, we keep the dataset size constant while increasing the number of IPUs. The reported performance metric, average throughput, is calculated as the number of graphs processed per second (similar to the spirit of evaluating the number of sequences processed per second in training language models). As can be observed from the figure, in most cases, increasing the number of IPUs increases the average throughput. With smaller datasets such as QM9, however, the throughput is maximized with 32 IPUs (with packing). Beyond 32 IPUs, the performance deteriorates due to having not enough work to sustain higher throughput. With the 2.7M and 4.5M water cluster datasets and with packing technique, the average throughput continues to increase up through 64 IPUs, as there are sufficient available work to keep all the independent processors busy (We also report the per-epoch MSE loss with 2.7M water cluster dataset on 16 IPUs in~\Cref{fig:per_epoch_loss} of the Appendix due to space constraint). 

\begin{figure}
    \centering
    \includegraphics[width=0.42\textwidth]{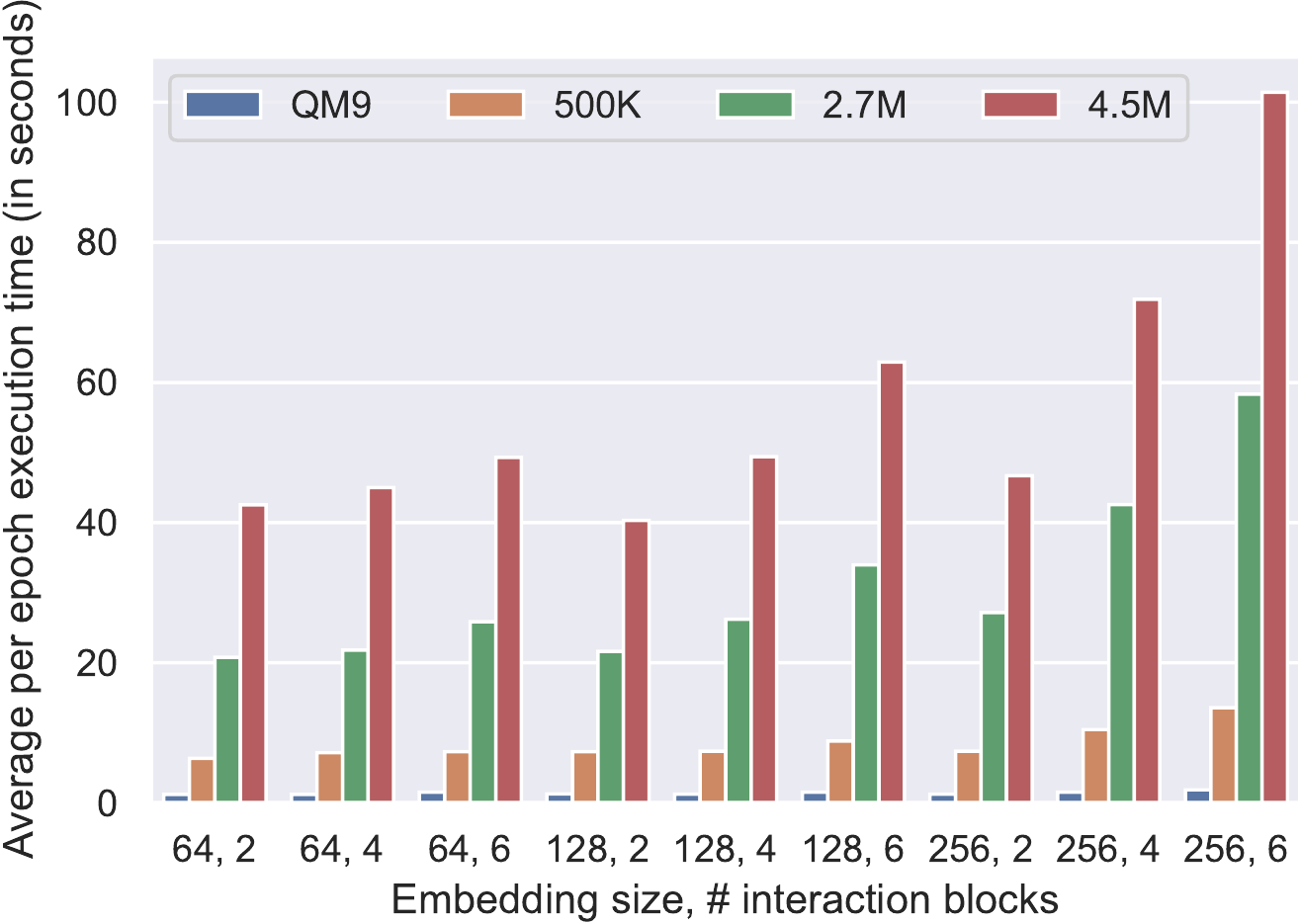} 
\vspace{-1.3em}    
    \caption{Performance of the SchNet model as the embedding size and the number of interaction blocks are varied.}
    \label{fig:embedding_interaction_block}
\vspace{-0.5em}
\end{figure}


\subsection{Impact of Embedding size and the Number of Interaction layers}
We also conducted experiments by varying the embedding size for each of the vertices of the input molecular graphs, as well as the number of interaction blocks in the SchNet model. We report the average per epoch execution time in ~\Cref{fig:embedding_interaction_block}. Except for the (64, 2) and (128,2) embedding size and \# interaction blocks combinations for the 4.5M dataset, as expected, the execution time increases with the increase of embedding size and the number of interaction blocks due to the increase in the total number of matrix multiplications performed in the SchNet model.

\subsection{Average Per Epoch Execution Time  
with Different Number of IPUs}
We also report the average per epoch execution time in seconds as we increase the number of IPUs in~\Cref{table:per_epoch_vary_ipu}.
With the QM9 dataset, the SchNet model achieves the best performance with 32 IPUs, closely followed by the performance while running on 16 IPUs. With QM9 dataset, on 64 IPUs, the overhead of communication starts to dominate the execution time instead of the available computation. With the 2.7M and 4.5M water cluster datasets, due to their larger sizes, increasing the number of IPUs help to achieve better performance as there are more local computation available.

\begin{table}[ht]
\small
\vspace{-0.5em}
\centering
\begin{tabular}{cccccc}
\toprule
Dataset       & 8 IPUs  & 16 IPUs & 32 IPUs & 64 IPUs & 8GPUs\\
\midrule
QM9  & 0.91 & 0.72 & 0.68 & 0.9 & 1.86\\
500K & 8.39 & 5.36 & 5.0 & 5.57 & 6.87 \\
2.7M & 35.07 & 21.37 & 14.81 & 11.74 & 34.36\\
4.5M & 62.56 & 35.0 & 27.03 & 19.38 & 60\\
\bottomrule
\end{tabular}
\vspace{0.5em}
\caption{\small \textit{Average per epoch execution time in seconds with different number of IPUs and GPUs.}
}\label{table:per_epoch_vary_ipu}
\vspace{-1.2em}
\end{table}

\subsection{Hardware accelerator comparison}
We compare the performance of training our optimized SchNet model for molecular GNNs on the IPUs with an existing out-of-the-box GPU implementation from the PyTorch Geometric (PyG) library. For running on multiple GPUs, we modified the existing single-GPU SchNet implementation in PyG to leverage the Distributed Data Parallel (DDP) model of the PyTorch library, without introducing any further optimizations. We conduct the experiments with different datasets for 25 epochs and report the result in~\Cref{table:per_epoch_vary_ipu}.
Here we report and compare the average per epoch execution time with 8 A100 GPUs and 16 IPUs because of their close equivalence in terms of compute capability, power and cost. 
As can be observed from~\Cref{table:per_epoch_vary_ipu}, 
the SchNet model runs 
faster on the IPUs compared to the GPUs. In particular, with QM9, 500k water cluster,  2.7M water cluster and 4.5M water cluster datasets, the SchNet model running on the IPUs achieve a speedup of 
2.58x, 1.28x, 1.6x and 1.71x respectively. The speedup is computed by taking the average of the execution time for 25 epochs for 16 IPUs and 8 GPUs 
then taking the ratio of the 
average runtime on the GPUs and on the IPUs. 

\section{Conclusion}
The long history of science is full of examples where new technology has led to a deeper understanding of natural phenomena.  More recently, progress in machine learning applied to computer vision and natural language processing has previously been driven by the re-purposing of GPU accelerators for the compute workloads of these deep neural networks \cite{AlexNet, NLPAttention}.  We present here the first application of IPUs to molecular GNNs and demonstrate extreme acceleration compared to a baseline GPU implementation. Beyond demonstrating this emerging capability, this paper proposes a set of optimization techniques for executing GNNs with molecular graphs: packing for efficient representation of the batches, planning for well-balanced work partitioning for optimized scatter-gather execution, overlapping computation and communication techniques for reducing the I/O bottleneck and model-specific optimizations. We demonstrate that altogether these optimizations can take advantage of the high-bandwidth local memory and faster interconnect of the IPU architecture to significantly improve the training time of the molecular GNNs. This has the potential to unlock the potential of machine learning applied to molecular-related scientific domains.

\clearpage
\pagebreak

\bibliography{reference}

\begin{thebibliography}{70}
\providecommand{\natexlab}[1]{#1}
\providecommand{\url}[1]{\texttt{#1}}
\expandafter\ifx\csname urlstyle\endcsname\relax
  \providecommand{\doi}[1]{doi: #1}\else
  \providecommand{\doi}{doi: \begingroup \urlstyle{rm}\Url}\fi

\bibitem[Addanki et~al.(2021)Addanki, Battaglia, Budden, Deac, Godwin, Keck,
  Li, Sanchez-Gonzalez, Stott, Thakoor, et~al.]{addanki2021large}
Addanki, R., Battaglia, P.~W., Budden, D., Deac, A., Godwin, J., Keck, T., Li,
  W. L.~S., Sanchez-Gonzalez, A., Stott, J., Thakoor, S., et~al.
\newblock Large-scale graph representation learning with very deep gnns and
  self-supervision.
\newblock \emph{arXiv preprint arXiv:2107.09422}, 2021.

\bibitem[Batzner et~al.(2022)Batzner, Musaelian, Sun, Geiger, Mailoa,
  Kornbluth, Molinari, Smidt, and Kozinsky]{batzner20223}
Batzner, S., Musaelian, A., Sun, L., Geiger, M., Mailoa, J.~P., Kornbluth, M.,
  Molinari, N., Smidt, T.~E., and Kozinsky, B.
\newblock E (3)-equivariant graph neural networks for data-efficient and
  accurate interatomic potentials.
\newblock \emph{Nature communications}, 13\penalty0 (1):\penalty0 1--11, 2022.

\bibitem[Behler(2017)]{behler2017first}
Behler, J.
\newblock First principles neural network potentials for reactive simulations
  of large molecular and condensed systems.
\newblock \emph{Angewandte Chemie International Edition}, 56\penalty0
  (42):\penalty0 12828--12840, 2017.

\bibitem[Bilbrey et~al.(2020)Bilbrey, Heindel, Schram, Bandyopadhyay, Xantheas,
  and Choudhury]{bilbrey2020jcp}
Bilbrey, J.~A., Heindel, J.~P., Schram, M., Bandyopadhyay, P., Xantheas, S.~S.,
  and Choudhury, S.
\newblock A look inside the black box: Using graph-theoretical descriptors to
  interpret a {Continuous-Filter Convolutional Neural Network} ({CF-CNN})
  trained on the global and local minimum energy structures of neutral water
  clusters.
\newblock \emph{The Journal of Chemical Physics}, 153\penalty0 (2):\penalty0
  024302, 2020.
\newblock \doi{10.1063/5.0009933}.

\bibitem[Cai et~al.(2021)Cai, Yan, Wu, Ma, Cheng, and Yu]{cai2021dgcl}
Cai, Z., Yan, X., Wu, Y., Ma, K., Cheng, J., and Yu, F.
\newblock Dgcl: An efficient communication library for distributed gnn
  training.
\newblock In \emph{Proceedings of the Sixteenth European Conference on Computer
  Systems}, pp.\  130--144, 2021.

\bibitem[Chanussot et~al.(2021)Chanussot, Das, Goyal, Lavril, Shuaibi, Riviere,
  Tran, Heras-Domingo, Ho, Hu, et~al.]{chanussot2021open}
Chanussot, L., Das, A., Goyal, S., Lavril, T., Shuaibi, M., Riviere, M., Tran,
  K., Heras-Domingo, J., Ho, C., Hu, W., et~al.
\newblock Open catalyst 2020 (oc20) dataset and community challenges.
\newblock \emph{ACS Catalysis}, 11\penalty0 (10):\penalty0 6059--6072, 2021.

\bibitem[Chiang et~al.(2019)Chiang, Liu, Si, Li, Bengio, and
  Hsieh]{chiang2019cluster}
Chiang, W.-L., Liu, X., Si, S., Li, Y., Bengio, S., and Hsieh, C.-J.
\newblock Cluster-gcn: An efficient algorithm for training deep and large graph
  convolutional networks.
\newblock In \emph{Proceedings of the 25th ACM International Conference on
  Knowledge Discovery \& Data Mining}, 2019.

\bibitem[Choudhury et~al.(2020{\natexlab{a}})Choudhury, Bilbrey, Ward,
  Xantheas, Foster, Heindel, Blaiszik, and Schwarting]{choudhury2020hydronet}
Choudhury, S., Bilbrey, J.~A., Ward, L., Xantheas, S.~S., Foster, I., Heindel,
  J.~P., Blaiszik, B., and Schwarting, M.~E.
\newblock Hydronet: Benchmark tasks for preserving intermolecular interactions
  and structural motifs in predictive and generative models for molecular data.
\newblock \emph{NeurIPS Workshop on Physical Sciences}, 2020{\natexlab{a}}.

\bibitem[Choudhury et~al.(2020{\natexlab{b}})Choudhury, Pope, Ward, Foster,
  Schwarting, Blaiszik, Heindel, and Xantheas]{HydroNetDataset}
Choudhury, S., Pope, J., Ward, L., Foster, I., Schwarting, M., Blaiszik, B.,
  Heindel, J., and Xantheas, S.
\newblock Hydronet: Benchmark tasks for preserving structural motifs and
  long-range interactions in predictive and generative models,
  2020{\natexlab{b}}.
\newblock URL \url{https://doi.org/10.18126/8PBB-YT6O}.

\bibitem[Emani et~al.(2021)Emani, Vishwanath, Adams, Papka, Stevens, Florescu,
  Jairath, Liu, Nama, and Sujeeth]{emani2021accelerating}
Emani, M., Vishwanath, V., Adams, C., Papka, M.~E., Stevens, R., Florescu, L.,
  Jairath, S., Liu, W., Nama, T., and Sujeeth, A.
\newblock Accelerating scientific applications with sambanova reconfigurable
  dataflow architecture.
\newblock \emph{Computing in Science \& Engineering}, 23\penalty0 (2):\penalty0
  114--119, 2021.

\bibitem[Fey \& Lenssen(2019{\natexlab{a}})Fey and Lenssen]{fey2019fast}
Fey, M. and Lenssen, J.~E.
\newblock Fast graph representation learning with pytorch geometric.
\newblock \emph{arXiv preprint arXiv:1903.02428}, 2019{\natexlab{a}}.

\bibitem[Fey \& Lenssen(2019{\natexlab{b}})Fey and Lenssen]{pyG}
Fey, M. and Lenssen, J.~E.
\newblock Fast graph representation learning with {PyTorch Geometric}.
\newblock In \emph{ICLR Workshop on Representation Learning on Graphs and
  Manifolds (ICLR)}, 2019{\natexlab{b}}.

\bibitem[Gandhi \& Iyer(2021)Gandhi and Iyer]{gandhi2021p3}
Gandhi, S. and Iyer, A.~P.
\newblock P3: Distributed deep graph learning at scale.
\newblock In \emph{15th $\{$USENIX$\}$ Symposium on Operating Systems Design
  and Implementation ($\{$OSDI$\}$ 21)}, pp.\  551--568, 2021.

\bibitem[Geng et~al.(2020)Geng, Li, Shi, Wu, Wang, Li, Haghi, Tumeo, Che,
  Reinhardt, et~al.]{geng2020awb}
Geng, T., Li, A., Shi, R., Wu, C., Wang, T., Li, Y., Haghi, P., Tumeo, A., Che,
  S., Reinhardt, S., et~al.
\newblock Awb-gcn: A graph convolutional network accelerator with runtime
  workload rebalancing.
\newblock In \emph{2020 53rd Annual IEEE/ACM International Symposium on
  Microarchitecture (MICRO)}, pp.\  922--936. IEEE, 2020.

\bibitem[Gilmer et~al.(2017)Gilmer, Schoenholz, Riley, Vinyals, and
  Dahl]{gilmer2017neural}
Gilmer, J., Schoenholz, S.~S., Riley, P.~F., Vinyals, O., and Dahl, G.~E.
\newblock Neural message passing for quantum chemistry.
\newblock \emph{arXiv preprint arXiv:1704.01212}, 2017.

\bibitem[Hamilton et~al.(2017)Hamilton, Ying, and Leskovec]{SageConv}
Hamilton, W., Ying, Z., and Leskovec, J.
\newblock Inductive representation learning on large graphs.
\newblock In \emph{Advances in neural information processing systems
  (NeurIPS)}, 2017.

\bibitem[Hosseini et~al.(2022)Hosseini, Simini, and
  Vishwanath]{hosseini2022operation}
Hosseini, R., Simini, F., and Vishwanath, V.
\newblock Operation-level performance benchmarking of graph neural networks for
  scientific applications.
\newblock \emph{arXiv preprint arXiv:2207.09955}, 2022.

\bibitem[Jha et~al.(2021)Jha, Gupta, Ward, Yang, Wolverton, Foster, Liao,
  Choudhary, and Agrawal]{jha2021enabling}
Jha, D., Gupta, V., Ward, L., Yang, Z., Wolverton, C., Foster, I., Liao, W.-k.,
  Choudhary, A., and Agrawal, A.
\newblock Enabling deeper learning on big data for materials informatics
  applications.
\newblock \emph{Scientific reports}, 11\penalty0 (1):\penalty0 1--12, 2021.

\bibitem[Jia et~al.(2019)Jia, Tillman, Maggioni, and
  Scarpazza]{jia2019dissecting}
Jia, Z., Tillman, B., Maggioni, M., and Scarpazza, D.~P.
\newblock Dissecting the graphcore ipu architecture via microbenchmarking.
\newblock \emph{arXiv preprint arXiv:1912.03413}, 2019.

\bibitem[Jia et~al.(2020)Jia, Lin, Gao, Zaharia, and Aiken]{jia2020improving}
Jia, Z., Lin, S., Gao, M., Zaharia, M., and Aiken, A.
\newblock Improving the accuracy, scalability, and performance of graph neural
  networks with roc.
\newblock \emph{Proceedings of Machine Learning and Systems}, 2:\penalty0
  187--198, 2020.

\bibitem[Johnson(1973)]{johnson1973near}
Johnson, D.~S.
\newblock \emph{Near-optimal bin packing algorithms}.
\newblock PhD thesis, Massachusetts Institute of Technology, 1973.

\bibitem[Johnson \& Garey(1985)Johnson and Garey]{Johnson1985}
Johnson, D.~S. and Garey, M.~R.
\newblock {A 7160 theorem for bin packing}.
\newblock \emph{Journal of Complexity}, 1\penalty0 (1):\penalty0 65--106, oct
  1985.
\newblock ISSN 0885064X.
\newblock \doi{10.1016/0885-064X(85)90022-6}.
\newblock URL
  \url{https://linkinghub.elsevier.com/retrieve/pii/0885064X85900226}.

\bibitem[Joshi et~al.(2021)Joshi, Gebauer, Bontha, Khazaieli, James, Brown, and
  Kumar]{joshi20213d}
Joshi, R.~P., Gebauer, N.~W., Bontha, M., Khazaieli, M., James, R.~M., Brown,
  J.~B., and Kumar, N.
\newblock 3d-scaffold: A deep learning framework to generate 3d coordinates of
  drug-like molecules with desired scaffolds.
\newblock \emph{The Journal of Physical Chemistry B}, 125\penalty0
  (44):\penalty0 12166--12176, 2021.

\bibitem[Jouppi et~al.(2017)Jouppi, Young, Patil, Patterson, Agrawal, Bajwa,
  Bates, Bhatia, Boden, Borchers, et~al.]{jouppi2017datacenter}
Jouppi, N.~P., Young, C., Patil, N., Patterson, D., Agrawal, G., Bajwa, R.,
  Bates, S., Bhatia, S., Boden, N., Borchers, A., et~al.
\newblock In-datacenter performance analysis of a tensor processing unit.
\newblock In \emph{Proceedings of the 44th annual international symposium on
  computer architecture}, pp.\  1--12, 2017.

\bibitem[Kaler et~al.(2022)Kaler, Stathas, Ouyang, Iliopoulos, Schardl,
  Leiserson, and Chen]{kaler2022accelerating}
Kaler, T., Stathas, N., Ouyang, A., Iliopoulos, A.-S., Schardl, T., Leiserson,
  C.~E., and Chen, J.
\newblock Accelerating training and inference of graph neural networks with
  fast sampling and pipelining.
\newblock \emph{Proceedings of Machine Learning and Systems}, 4:\penalty0
  172--189, 2022.

\bibitem[Karypis \& Kumar(1998)Karypis and Kumar]{karypis1998multilevelk}
Karypis, G. and Kumar, V.
\newblock Multilevelk-way partitioning scheme for irregular graphs.
\newblock \emph{Journal of Parallel and Distributed computing}, 48\penalty0
  (1):\penalty0 96--129, 1998.

\bibitem[Kipf \& Welling(2017)Kipf and Welling]{GCNConv}
Kipf, T.~N. and Welling, M.
\newblock Semi-supervised classification with graph convolutional networks.
\newblock \emph{International Conference on Learning Representations (ICLR)},
  2017.

\bibitem[Klicpera et~al.(2020)Klicpera, Gro{\ss}, and
  G{\"u}nnemann]{klicpera2020directional}
Klicpera, J., Gro{\ss}, J., and G{\"u}nnemann, S.
\newblock Directional message passing for molecular graphs.
\newblock \emph{arXiv preprint arXiv:2003.03123}, 2020.

\bibitem[Kohn(1996)]{KohnPhysRevLett.76.3168}
Kohn, W.
\newblock Density functional and density matrix method scaling linearly with
  the number of atoms.
\newblock \emph{Phys. Rev. Lett.}, 76:\penalty0 3168--3171, Apr 1996.
\newblock \doi{10.1103/PhysRevLett.76.3168}.
\newblock URL \url{https://link.aps.org/doi/10.1103/PhysRevLett.76.3168}.

\bibitem[Kohn \& Sham(1965)Kohn and Sham]{KohnShamPhysRev.140.A1133}
Kohn, W. and Sham, L.~J.
\newblock Self-consistent equations including exchange and correlation effects.
\newblock \emph{Phys. Rev.}, 140:\penalty0 A1133--A1138, Nov 1965.
\newblock \doi{10.1103/PhysRev.140.A1133}.
\newblock URL \url{https://link.aps.org/doi/10.1103/PhysRev.140.A1133}.

\bibitem[Korte \& Vygen(2012)Korte and Vygen]{Korte2012}
Korte, B. and Vygen, J.
\newblock \emph{{Combinatorial Optimization}}, volume~21 of \emph{Algorithms
  and Combinatorics}.
\newblock Springer Berlin Heidelberg, Berlin, Heidelberg, 2012.
\newblock ISBN 978-3-642-24487-2.
\newblock \doi{10.1007/978-3-642-24488-9}.
\newblock URL \url{http://link.springer.com/10.1007/978-3-642-24488-9}.

\bibitem[Krell et~al.(2021)Krell, Kosec, Perez, and Fitzgibbon]{packing}
Krell, M.~M., Kosec, M., Perez, S.~P., and Fitzgibbon, A.
\newblock Efficient sequence packing without cross-contamination: Accelerating
  large language models without impacting performance, 2021.
\newblock URL \url{https://arxiv.org/abs/2107.02027}.

\bibitem[Krizhevsky et~al.(2012)Krizhevsky, Sutskever, and Hinton]{AlexNet}
Krizhevsky, A., Sutskever, I., and Hinton, G.~E.
\newblock Imagenet classification with deep convolutional neural networks.
\newblock In Pereira, F., Burges, C., Bottou, L., and Weinberger, K. (eds.),
  \emph{Advances in Neural Information Processing Systems}, volume~25. Curran
  Associates, Inc., 2012.
\newblock URL
  \url{https://proceedings.neurips.cc/paper/2012/file/c399862d3b9d6b76c8436e924a68c45b-Paper.pdf}.

\bibitem[Kulichenko et~al.(2021)Kulichenko, Smith, Nebgen, Li, Fedik, Boldyrev,
  Lubbers, Barros, and Tretiak]{kulichenko2021rise}
Kulichenko, M., Smith, J.~S., Nebgen, B., Li, Y.~W., Fedik, N., Boldyrev,
  A.~I., Lubbers, N., Barros, K., and Tretiak, S.
\newblock The rise of neural networks for materials and chemical dynamics.
\newblock \emph{The Journal of Physical Chemistry Letters}, 12\penalty0
  (26):\penalty0 6227--6243, 2021.

\bibitem[Lee \& Lee(1985)Lee and Lee]{Lee1985}
Lee, C.~C. and Lee, D.~T.
\newblock {A Simple On-Line Bin-Packing Algorithm}.
\newblock \emph{Journal of the ACM (JACM)}, 32\penalty0 (3):\penalty0 562--572,
  jul 1985.
\newblock ISSN 1557735X.
\newblock \doi{10.1145/3828.3833}.
\newblock URL \url{https://dl.acm.org/doi/10.1145/3828.3833}.

\bibitem[Lim et~al.(2019)Lim, Ryu, Park, Choe, Ham, and Kim]{lim2019predicting}
Lim, J., Ryu, S., Park, K., Choe, Y.~J., Ham, J., and Kim, W.~Y.
\newblock Predicting drug--target interaction using a novel graph neural
  network with 3d structure-embedded graph representation.
\newblock \emph{Journal of chemical information and modeling}, 59\penalty0
  (9):\penalty0 3981--3988, 2019.

\bibitem[Lubbers et~al.(2018)Lubbers, Smith, and Barros]{HIPNN}
Lubbers, N., Smith, J.~S., and Barros, K.
\newblock Hierarchical modeling of molecular energies using a deep neural
  network.
\newblock \emph{The Journal of Chemical Physics}, 148\penalty0 (24):\penalty0
  241715, 2018.
\newblock \doi{10.1063/1.5011181}.
\newblock URL \url{https://doi.org/10.1063/1.5011181}.

\bibitem[Ma et~al.(2019)Ma, Yang, Miao, Xue, Wu, Zhou, and Dai]{ma2019neugraph}
Ma, L., Yang, Z., Miao, Y., Xue, J., Wu, M., Zhou, L., and Dai, Y.
\newblock Neugraph: parallel deep neural network computation on large graphs.
\newblock In \emph{USENIX Annual Technical Conference}, 2019.

\bibitem[Moe et~al.(2022)Moe, Pogorelov, Schroeder, and
  Langguth]{moe2022implementating}
Moe, J., Pogorelov, K., Schroeder, D.~T., and Langguth, J.
\newblock Implementating spatio-temporal graph convolutional networks on
  graphcore ipus.
\newblock In \emph{2022 IEEE International Parallel and Distributed Processing
  Symposium Workshops (IPDPSW)}, pp.\  45--54. IEEE, 2022.

\bibitem[Morari et~al.(2014)Morari, Tumeo, Chavarr{\'\i}a-Miranda, Villa, and
  Valero]{morari2014scaling}
Morari, A., Tumeo, A., Chavarr{\'\i}a-Miranda, D., Villa, O., and Valero, M.
\newblock Scaling irregular applications through data aggregation and software
  multithreading.
\newblock In \emph{2014 IEEE 28th International Parallel and Distributed
  Processing Symposium}, pp.\  1126--1135. IEEE, 2014.

\bibitem[Pinheiro et~al.(2020)Pinheiro, Mucelini, Soares, Prati, Da~Silva, and
  Quiles]{pinheiro2020machine}
Pinheiro, G.~A., Mucelini, J., Soares, M.~D., Prati, R.~C., Da~Silva, J.~L.,
  and Quiles, M.~G.
\newblock Machine learning prediction of nine molecular properties based on the
  smiles representation of the qm9 quantum-chemistry dataset.
\newblock \emph{The Journal of Physical Chemistry A}, 124\penalty0
  (47):\penalty0 9854--9866, 2020.

\bibitem[PopTorch(2022)]{Poptorch_git}
PopTorch, G.
\newblock Poptorch: Pytorch integration for the graphcore ipu, 2022.
\newblock URL \url{https://github.com/graphcore/poptorch}.

\bibitem[Prentice et~al.(2020)Prentice, Aarons, and
  Womack]{prentice_onetep_2020}
Prentice, J. C.~A., Aarons, J., and Womack, J.~C.
\newblock The {ONETEP} linear-scaling density functional theory program.
\newblock \emph{J. Chem. Phys}, 152:\penalty0 174111, 2020.
\newblock \doi{10.1063/5.0004445}.
\newblock URL \url{https://doi.org/10.1063/5.0004445}.

\bibitem[Prodan \& Kohn(2005)Prodan and Kohn]{Prodan_2005}
Prodan, E. and Kohn, W.
\newblock Nearsightedness of electronic matter.
\newblock \emph{Proceedings of the National Academy of Sciences}, 102\penalty0
  (33):\penalty0 11635--11638, aug 2005.
\newblock \doi{10.1073/pnas.0505436102}.
\newblock URL \url{https://doi.org/10.1073%2Fpnas.0505436102}.

\bibitem[Rakshit et~al.(2019)Rakshit, Bandyopadhyay, Heindel, and
  Xantheas]{rakshit2019atlas}
Rakshit, A., Bandyopadhyay, P., Heindel, J.~P., and Xantheas, S.~S.
\newblock Atlas of putative minima and low-lying energy networks of water
  clusters n= 3--25.
\newblock \emph{The Journal of chemical physics}, 151\penalty0 (21):\penalty0
  214307, 2019.

\bibitem[Ramakrishnan et~al.(2014)Ramakrishnan, Dral, Rupp, and von
  Lilienfeld]{ramakrishnan2014quantum}
Ramakrishnan, R., Dral, P.~O., Rupp, M., and von Lilienfeld, O.~A.
\newblock Quantum chemistry structures and properties of 134 kilo molecules.
\newblock \emph{Scientific Data}, 1, 2014.

\bibitem[Schmidt et~al.(2019)Schmidt, Marques, Botti, and
  Marques]{schmidt2019recent}
Schmidt, J., Marques, M.~R., Botti, S., and Marques, M.~A.
\newblock Recent advances and applications of machine learning in solid-state
  materials science.
\newblock \emph{npj Computational Materials}, 5\penalty0 (1):\penalty0 1--36,
  2019.

\bibitem[Schwaller et~al.(2019)Schwaller, Laino, Gaudin, Bolgar, Hunter, Bekas,
  and Lee]{schwaller2019molecular}
Schwaller, P., Laino, T., Gaudin, T., Bolgar, P., Hunter, C.~A., Bekas, C., and
  Lee, A.~A.
\newblock Molecular transformer: a model for uncertainty-calibrated chemical
  reaction prediction.
\newblock \emph{ACS central science}, 5\penalty0 (9):\penalty0 1572--1583,
  2019.

\bibitem[Schütt et~al.(2017)Schütt, Arbabzadah, Chmiela, Müller, and
  Tkatchenko]{DTNN2017}
Schütt, K.~T., Arbabzadah, F., Chmiela, S., Müller, K.~R., and Tkatchenko, A.
\newblock Quantum-chemical insights from deep tensor neural networks.
\newblock \emph{Nature Communications}, 8\penalty0 (1):\penalty0 13890, 2017.
\newblock ISSN 2041-1723.
\newblock \doi{10.1038/ncomms13890}.
\newblock URL \url{https://doi.org/10.1038/ncomms13890}.

\bibitem[Schütt et~al.(2018)Schütt, Sauceda, Kindermans, Tkatchenko, and
  Müller]{SchNet2018}
Schütt, K.~T., Sauceda, H.~E., Kindermans, P.-J., Tkatchenko, A., and Müller,
  K.-R.
\newblock Schnet – a deep learning architecture for molecules and materials.
\newblock \emph{The Journal of Chemical Physics}, 148\penalty0 (24):\penalty0
  241722, 2018.
\newblock \doi{10.1063/1.5019779}.

\bibitem[Schütt et~al.(2019)Schütt, Kessel, Gastegger, Nicoli, Tkatchenko,
  and Müller]{SchNetPack2019}
Schütt, K.~T., Kessel, P., Gastegger, M., Nicoli, K.~A., Tkatchenko, A., and
  Müller, K.-R.
\newblock Schnetpack: A deep learning toolbox for atomistic systems.
\newblock \emph{Journal of Chemical Theory and Computation}, 15\penalty0
  (1):\penalty0 448--455, 2019.
\newblock \doi{10.1021/acs.jctc.8b00908}.

\bibitem[Smith et~al.(2017)Smith, Isayev, and Roitberg]{Smith2017}
Smith, J.~S., Isayev, O., and Roitberg, A.~E.
\newblock Ani-1: an extensible neural network potential with dft accuracy at
  force field computational cost.
\newblock \emph{Chem. Sci.}, 8:\penalty0 3192--3203, 2017.
\newblock \doi{10.1039/C6SC05720A}.
\newblock URL \url{http://dx.doi.org/10.1039/C6SC05720A}.

\bibitem[Sussman et~al.(1998)Sussman, Lin, Jiang, Manning, Prilusky, Ritter,
  and Abola]{sussman1998protein}
Sussman, J.~L., Lin, D., Jiang, J., Manning, N.~O., Prilusky, J., Ritter, O.,
  and Abola, E.~E.
\newblock Protein data bank (pdb): database of three-dimensional structural
  information of biological macromolecules.
\newblock \emph{Acta Crystallographica Section D: Biological Crystallography},
  54\penalty0 (6):\penalty0 1078--1084, 1998.

\bibitem[Tripathy et~al.(2020)Tripathy, Yelick, and
  Bulu{\c{c}}]{tripathy2020reducing}
Tripathy, A., Yelick, K., and Bulu{\c{c}}, A.
\newblock Reducing communication in graph neural network training.
\newblock In \emph{SC20: International Conference for High Performance
  Computing, Networking, Storage and Analysis}, pp.\  1--14. IEEE, 2020.

\bibitem[Vaswani et~al.(2017)Vaswani, Shazeer, Parmar, Uszkoreit, Jones, Gomez,
  Kaiser, and Polosukhin]{NLPAttention}
Vaswani, A., Shazeer, N., Parmar, N., Uszkoreit, J., Jones, L., Gomez, A.~N.,
  Kaiser, L., and Polosukhin, I.
\newblock Attention is all you need, 2017.
\newblock URL \url{https://arxiv.org/abs/1706.03762}.

\bibitem[Veličković et~al.(2018)Veličković, Cucurull, Casanova, Romero,
  Liò, and Bengio]{GATConv}
Veličković, P., Cucurull, G., Casanova, A., Romero, A., Liò, P., and Bengio,
  Y.
\newblock Graph attention networks.
\newblock In \emph{International Conference on Learning Representations
  (ICLR)}, 2018.

\bibitem[Wan et~al.(2022)Wan, Li, Wolfe, Kyrillidis, Kim, and
  Lin]{wan2022pipegcn}
Wan, C., Li, Y., Wolfe, C.~R., Kyrillidis, A., Kim, N.~S., and Lin, Y.
\newblock Pipegcn: Efficient full-graph training of graph convolutional
  networks with pipelined feature communication.
\newblock In \emph{International Conference on Learning Representations}, 2022.

\bibitem[Wang et~al.(2019)Wang, Yu, Zheng, Gan, Gai, Ye, Li, Zhou, Huang, Ma,
  Huang, Guo, Zhang, Lin, Zhao, Li, Smola, and Zhang]{wang2019dgl}
Wang, M., Yu, L., Zheng, D., Gan, Q., Gai, Y., Ye, Z., Li, M., Zhou, J., Huang,
  Q., Ma, C., Huang, Z., Guo, Q., Zhang, H., Lin, H., Zhao, J., Li, J., Smola,
  A.~J., and Zhang, Z.
\newblock Deep graph library: Towards efficient and scalable deep learning on
  graphs.
\newblock \emph{ICLR Workshop on Representation Learning on Graphs and
  Manifolds}, 2019.

\bibitem[Wang et~al.(2017)Wang, Pan, Davidson, Wu, Yang, Wang, Osama, Yuan,
  Liu, Riffel, et~al.]{wang2017gunrock}
Wang, Y., Pan, Y., Davidson, A., Wu, Y., Yang, C., Wang, L., Osama, M., Yuan,
  C., Liu, W., Riffel, A.~T., et~al.
\newblock Gunrock: Gpu graph analytics.
\newblock \emph{ACM Transactions on Parallel Computing (TOPC)}, 4\penalty0
  (1):\penalty0 1--49, 2017.

\bibitem[Wang et~al.(2021{\natexlab{a}})Wang, Feng, and Ding]{wang2021tc}
Wang, Y., Feng, B., and Ding, Y.
\newblock Tc-gnn: Accelerating sparse graph neural network computation via
  dense tensor core on gpus.
\newblock \emph{arXiv preprint arXiv:2112.02052}, 2021{\natexlab{a}}.

\bibitem[Wang et~al.(2021{\natexlab{b}})Wang, Feng, Li, Li, Deng, Xie, and
  Ding]{GNNAdvisor}
Wang, Y., Feng, B., Li, G., Li, S., Deng, L., Xie, Y., and Ding, Y.
\newblock Gnnadvisor: An efficient runtime system for gnn acceleration on gpus.
\newblock In \emph{USENIX Symposium on Operating Systems Design and
  Implementation (OSDI'21)}, 2021{\natexlab{b}}.

\bibitem[Wang et~al.(2021{\natexlab{c}})Wang, Feng, Li, Li, Deng, Xie, and
  Ding]{wang2021gnnadvisor}
Wang, Y., Feng, B., Li, G., Li, S., Deng, L., Xie, Y., and Ding, Y.
\newblock $\{$GNNAdvisor$\}$: An adaptive and efficient runtime system for
  $\{$GNN$\}$ acceleration on $\{$GPUs$\}$.
\newblock In \emph{15th USENIX Symposium on Operating Systems Design and
  Implementation (OSDI 21)}, pp.\  515--531, 2021{\natexlab{c}}.

\bibitem[Wang et~al.(2022)Wang, Wang, Cao, and
  Barati~Farimani]{wang2022molecular}
Wang, Y., Wang, J., Cao, Z., and Barati~Farimani, A.
\newblock Molecular contrastive learning of representations via graph neural
  networks.
\newblock \emph{Nature Machine Intelligence}, 4\penalty0 (3):\penalty0
  279--287, 2022.

\bibitem[Xu et~al.(2019)Xu, Hu, Leskovec, and Jegelka]{GINConv}
Xu, K., Hu, W., Leskovec, J., and Jegelka, S.
\newblock How powerful are graph neural networks?
\newblock In \emph{International Conference on Learning Representations
  (ICLR)}, 2019.

\bibitem[Yue \& Zhang(1995)Yue and Zhang]{Yue1995}
Yue, M. and Zhang, L.
\newblock {A simple proof of the inequality $MFFD(L)\leq71/60 OPT(L) + 1,L$ for
  the MFFD bin-packing algorithm}.
\newblock \emph{Acta Mathematicae Applicatae Sinica}, 11\penalty0 (3):\penalty0
  318--330, jul 1995.
\newblock ISSN 01689673.
\newblock \doi{10.1007/BF02011198}.

\bibitem[Zeng et~al.(2019)Zeng, Zhou, Srivastava, Kannan, and
  Prasanna]{zeng2019graphsaint}
Zeng, H., Zhou, H., Srivastava, A., Kannan, R., and Prasanna, V.
\newblock Graphsaint: Graph sampling based inductive learning method.
\newblock In \emph{International Conference on Learning Representations}, 2019.

\bibitem[Zhang et~al.(2020)Zhang, Liu, and Xie]{zhang2020molecular}
Zhang, S., Liu, Y., and Xie, L.
\newblock Molecular mechanics-driven graph neural network with multiplex graph
  for molecular structures.
\newblock \emph{arXiv preprint arXiv:2011.07457}, 2020.

\bibitem[Zheng et~al.(2020{\natexlab{a}})Zheng, Ma, Wang, Zhou, Su, Song, Gan,
  Zhang, and Karypis]{zheng2020distdgl}
Zheng, D., Ma, C., Wang, M., Zhou, J., Su, Q., Song, X., Gan, Q., Zhang, Z.,
  and Karypis, G.
\newblock Distdgl: distributed graph neural network training for billion-scale
  graphs.
\newblock In \emph{2020 IEEE/ACM 10th Workshop on Irregular Applications:
  Architectures and Algorithms (IA3)}, pp.\  36--44. IEEE, 2020{\natexlab{a}}.

\bibitem[Zheng et~al.(2020{\natexlab{b}})Zheng, Song, Ma, Tan, Ye, Dong, Xiong,
  Zhang, and Karypis]{zheng2020dgl}
Zheng, D., Song, X., Ma, C., Tan, Z., Ye, Z., Dong, J., Xiong, H., Zhang, Z.,
  and Karypis, G.
\newblock Dgl-ke: Training knowledge graph embeddings at scale.
\newblock In \emph{Proceedings of the 43rd International ACM SIGIR Conference
  on Research and Development in Information Retrieval}, pp.\  739--748,
  2020{\natexlab{b}}.

\bibitem[Zheng et~al.(2022)Zheng, Song, Yang, LaSalle, and
  Karypis]{zheng2022distributed}
Zheng, D., Song, X., Yang, C., LaSalle, D., and Karypis, G.
\newblock Distributed hybrid cpu and gpu training for graph neural networks on
  billion-scale heterogeneous graphs.
\newblock In \emph{Proceedings of the 28th ACM SIGKDD Conference on Knowledge
  Discovery and Data Mining}, pp.\  4582--4591, 2022.

\end{thebibliography}
\bibliographystyle{mlsys2023}


\clearpage
\pagebreak

\appendix
\section{Appendix}

\subsection{Additional Experimental results}

\subsubsection{Per Epoch MSE Loss for SchNet IPU}
\begin{figure}[htbp]
  \includegraphics[width=0.45\textwidth]{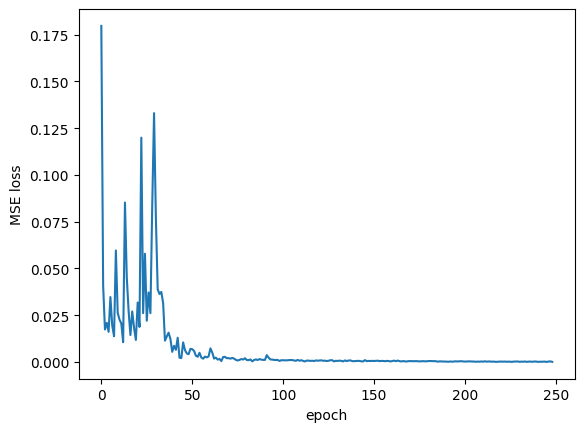}
  \caption{Per epoch MSE loss for the SchNet model with 2.7M water cluster dataset.} \label{fig:per_epoch_loss}
\end{figure}

We report per epoch MSE loss for the SchNet model with 2.7M water cluster dataset in~\Cref{fig:per_epoch_loss}. 

\subsubsection{Profiler Output to Demonstrate the Effect of Merging Communication Collectives}
\begin{figure*}[htbp]
  \includegraphics[width=\textwidth]{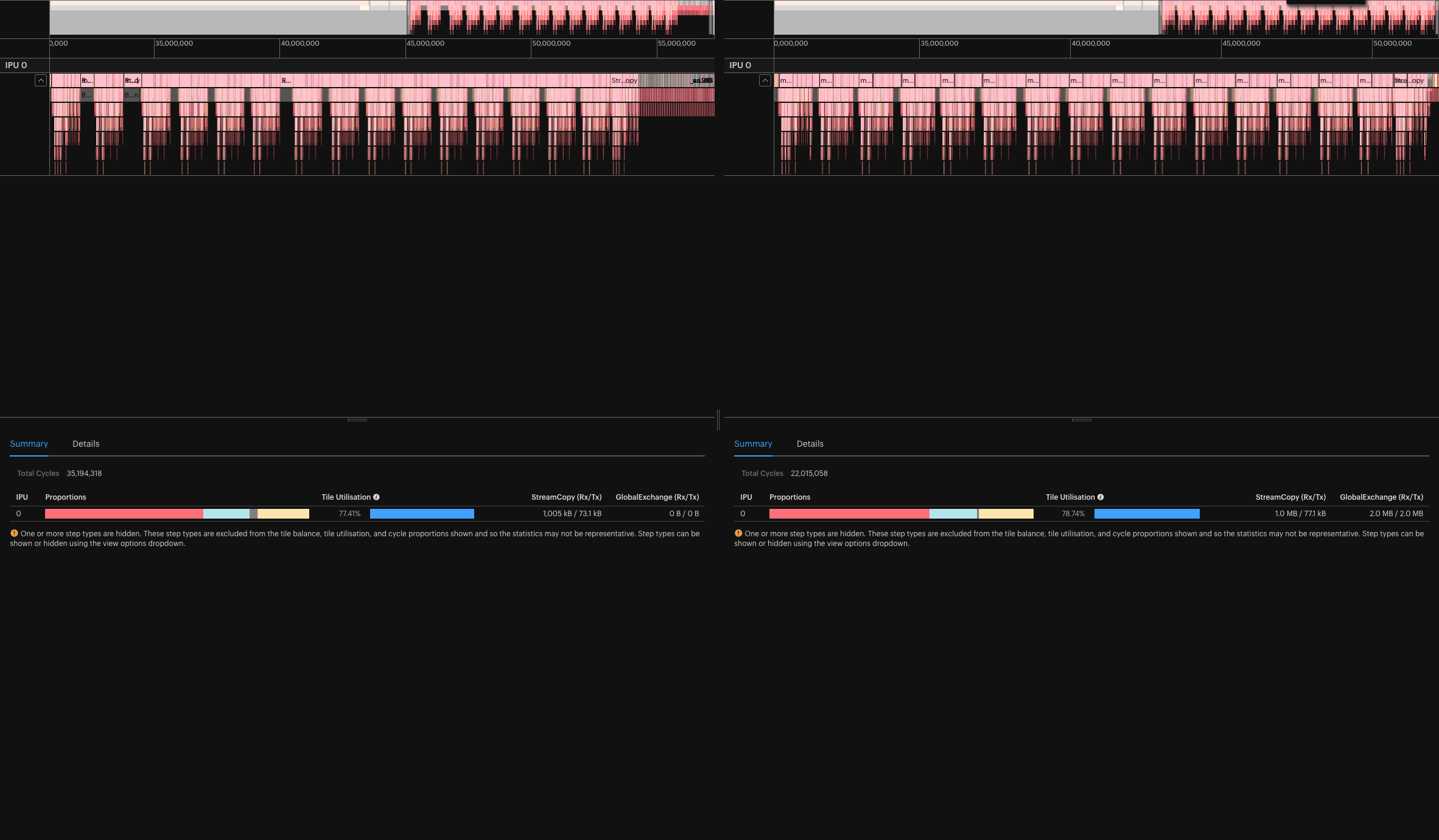}
  \caption{Effect of merging of all the allreduce collectives at the end of the backward pass to reduce the tail latency of the execution. On the left profile, from at around 52k time unit mark till the end, only some of the tiles are busy while other tiles are waiting. In contrast, on the right, till the very end of the execution, most of the tiles are engaged reasonably in computation and communication without long waiting period, thanks to the merging of the all reductions.} \label{fig:merge_allreduce_profile}
\end{figure*}

We show the effect of merging communication collectives for weight updates (discussed in ~\Cref{sec:model_specific_optim}) by profiling with and without this optimization in~\Cref{fig:merge_allreduce_profile}.

\subsubsection{Per Epoch Execution time With Different Number of IPUs}
\begin{figure*}
  \centering
  \begin{subfigure}[b]{0.33\textwidth}
  \centering
  \includegraphics[width=\textwidth]{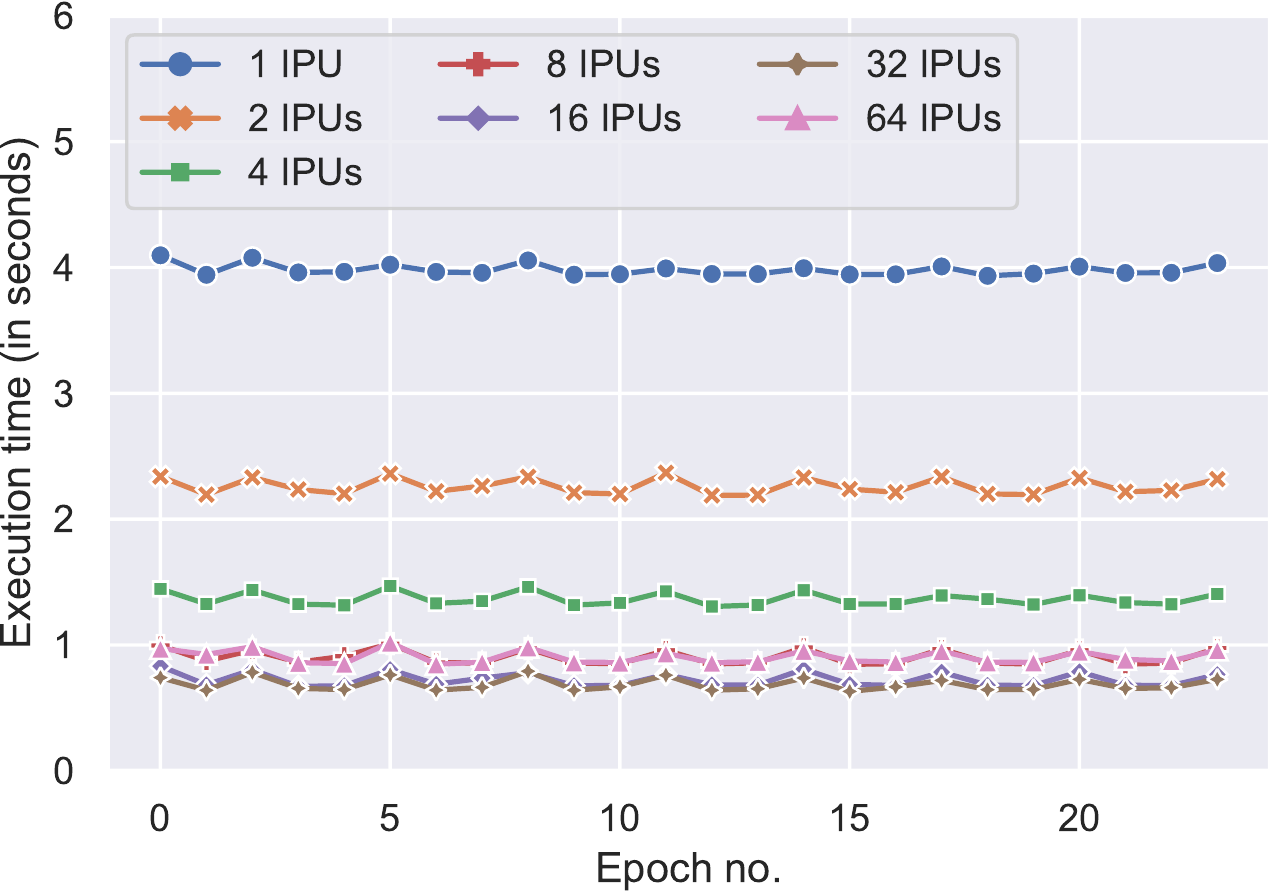}
  \caption{QM9}\label{fig:ipu_scaling_QM9}
  \end{subfigure} 
    \begin{subfigure}[b]{0.33\textwidth}
  \centering
  \includegraphics[width=\textwidth]{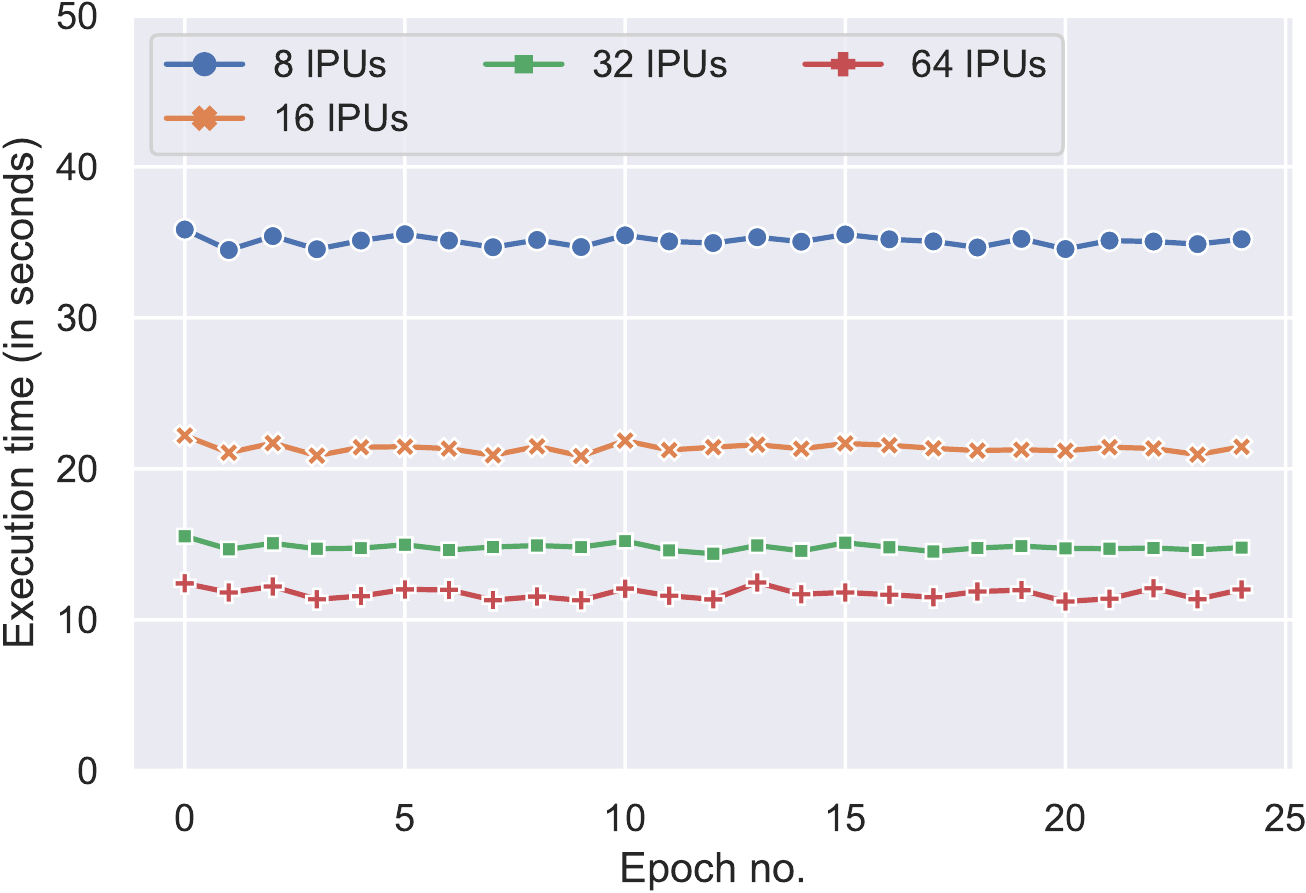}
  \caption{2.7M water cluster}\label{fig:ipu_scaling_2M}
  \end{subfigure}
    \begin{subfigure}[b]{0.33\textwidth}
  \centering
  \includegraphics[width=\textwidth]{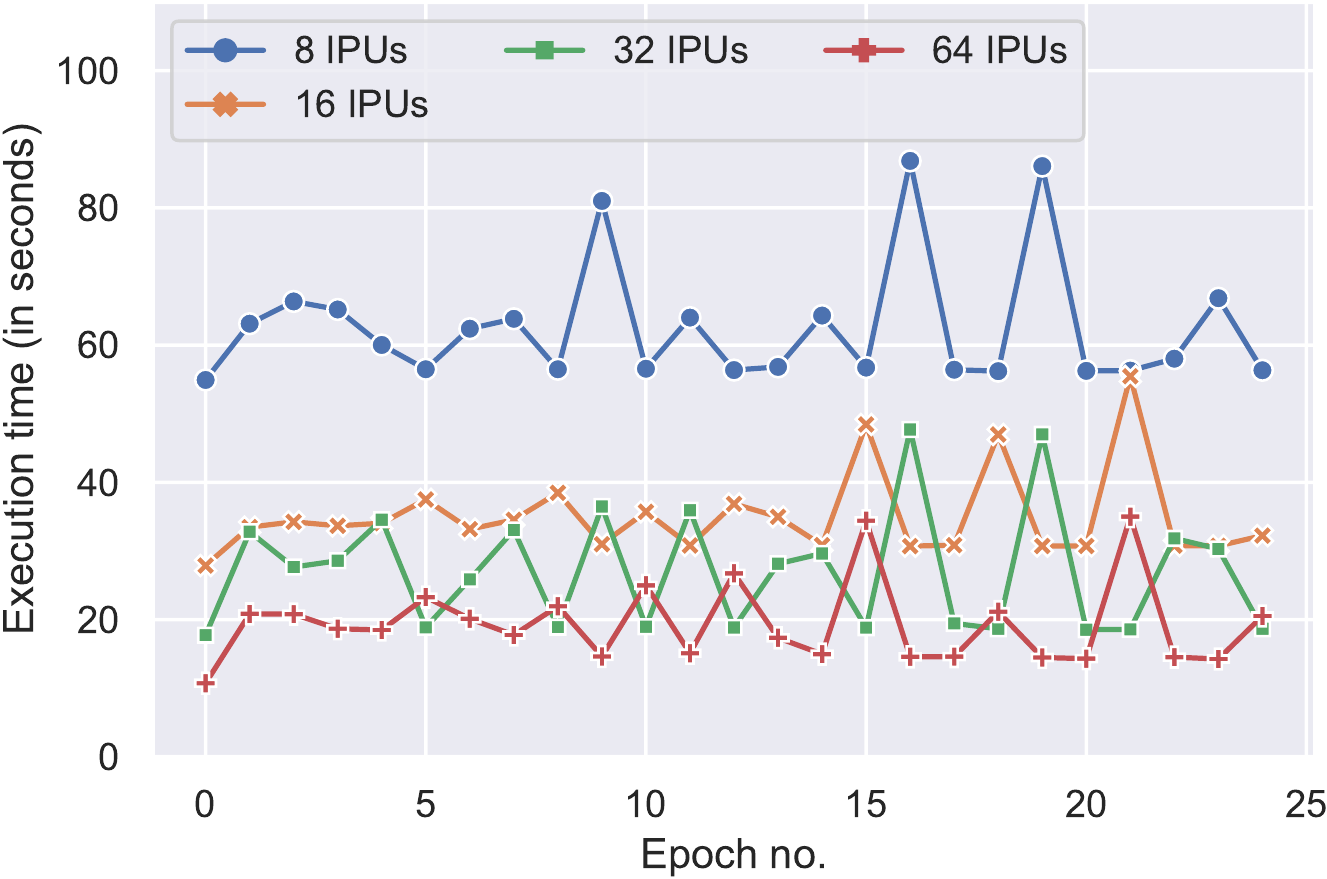}
   \caption{4.5M water cluster}\label{fig:ipu_scaling_5M}
  \end{subfigure}
\caption{Per epoch performance of the SchNet model with different number of IPUs and with packing (25 epochs).}\label{fig:ipu_scaling}
\end{figure*}

We report per epoch execution time while varying the number of IPUs in~\Cref{fig:ipu_scaling}.

\subsection{Extended Discussion on Related Work}
A plethora of frameworks has been proposed in the literature that seek to improve the performance of the GNNs~\cite{jia2020improving,zheng2020distdgl,pyG,cai2021dgcl,ma2019neugraph,wang2021gnnadvisor}. Among these frameworks,
Pytorch Geometric (PyG)~\cite{pyG} is the most well-known library for GNNs and leverages the torch-scatter library for graph aggregation operation. A recent work by Hosseini et al.\cite{hosseini2022operation} reported profiling results for the low-level operations in PyG and reported that, on the GPUs, memory is the main bottleneck in GNN execution. Deep Graph Library (DGL)~\cite{wang2019dgl} is another popular GNN library that uses the Sparse Matrix multiplication (SpMM) kernels from CuSparse for sum-reduction aggregation. NeuGraph~\cite{ma2019neugraph} introduced the "Scatter-ApplyEdge-Gather-ApplyVertex" abstraction for GNN computation. Kaler et al.~\cite{kaler2022accelerating} proposed customized neighborhood sampler, shared-memory parallelization and pipelining of batch transfer with GPU computation for improving GPU utilization. Other pipelining techniques for overlapping computation with communication have also been suggested~\cite{wan2022pipegcn}.


\subsubsection{GPU-related Optimizations for GNNs}
To date, both algorithmic and system-level optimizations for GNNs have been mostly geared towards Graphic Processing Units (GPUs) ~\cite{GCNConv,GINConv,ma2019neugraph,wang2019dgl,pyG} due to the availability of the GPUs (Nvidia A100, AMD MI200, etc.) on many current HPC systems. At the algorithm level, neighborhood sampling and mini batch processing have been proposed for reducing communication~\cite{chiang2019cluster,zeng2019graphsaint,gandhi2021p3}. Alternatively, techniques including pipelined batch transfer and computation~\cite{kaler2022accelerating}, specialized runtime and communication libraries~\cite{cai2021dgcl,ma2019neugraph,wang2021gnnadvisor} seek to improve the performance of the GNN libraries at the system level. Moreover, specialized hardware, such as tensor cores inside the GPUs  have been adapted to accelerate the sparse-dense matrix multiplication step in the aggregation phase of the GNNs~\cite{wang2021tc}. 
GPUs primarily rely on the oversubscription of threads to hide memory latency as well as coalesced memory access within each warp to achieve peak memory bandwidth. For these reasons, GPUs perform well when the computation is regular and dense. However, fine-grained, irregular workloads such as GNNs require significant engineering to obtain optimal performance on the GPUs~\cite{wang2017gunrock}.

\subsubsection{IPUs and ML Workload}
For and overview of the IPU hardware and various benchmarking results on these systems, please see the white-paper ~\cite{jia2019dissecting}. Discussion about an implementation of Spatio-Temporal Graph Convolutional Networks on Graphcore IPUs can be found in~\cite{moe2022implementating}.




\end{document}